\documentclass[final,preprint]{elsarticle}

\usepackage{lineno}
\usepackage{xspace}

\modulolinenumbers[5] 
\usepackage{multirow} 
\usepackage{booktabs}
\usepackage{makecell}
\usepackage{bbding}
\usepackage{graphicx}
\usepackage{comment}
\usepackage{amsmath,amssymb} 
\usepackage{ulem}
\usepackage{enumitem}
\usepackage{amsthm}
\usepackage{subfig}
\usepackage{algorithm}
\usepackage{algorithmic}
\usepackage{url}
\usepackage{hyperref}
\usepackage[hyphenbreaks]{breakurl}
\usepackage {color}

\newtheorem{proposition}{Proposition}

\pdfoutput=1

\begin{document}

\begin{frontmatter}

\title{One-shot Federated Learning without Server-side Training}

\author{Shangchao Su\fnref{myfootnote1}}
\author{Bin Li\fnref{myfootnote1}\corref{mycorrespondingauthor} }
\author{Xiangyang Xue\fnref{myfootnote1}} 
\address{School of Computer Science, Fudan University, China}
\fntext[myfootnote1]{S. Su (e-mail: scsu20@fudan.edu.cn), B. Li (e-mail: libin@fudan.edu.cn) and X. Xue (e-mail: xyxue@fudan.edu.cn) are with the Shanghai Key Laboratory of Intelligent Information Processing and School of Computer Science, Fudan University. }
\cortext[mycorrespondingauthor]{Corresponding author}

\begin{abstract}
 Federated Learning (FL) has recently made significant progress as a new machine learning paradigm for privacy protection. Due to the high communication cost of traditional FL, one-shot federated learning is gaining popularity as a way to reduce communication cost between clients and the server. Most of the existing one-shot FL methods are based on Knowledge Distillation; however, {distillation based approach requires an extra training phase and depends on publicly available data sets or generated pseudo samples.} In this work, we consider a novel and challenging cross-silo setting: performing a single round of parameter aggregation on the local models without server-side training. In this setting, we propose an effective algorithm for Model Aggregation via Exploring Common Harmonized Optima (MA-Echo), which iteratively updates the parameters of all local models to bring them close to a common low-loss area on the loss surface, without harming performance on their own data sets at the same time. Compared to the existing methods, MA-Echo can work well even in extremely non-identical data distribution settings where the support categories of each local model have no overlapped labels with those of the others. We conduct extensive experiments on two popular image classification data sets to compare the proposed method with existing methods and demonstrate the effectiveness of MA-Echo, which clearly outperforms the state-of-the-arts. The source code can be accessed in \url{https://github.com/FudanVI/MAEcho}.
 \end{abstract}


\end{frontmatter}


\section{Introduction}
With their powerful representation capabilities, deep neural networks have achieved great success in various learning tasks such as image and text classification~\cite{krizhevsky2012imagenet,he2016deep,kim2014convolutional}. However, in many real-world application scenarios such as multi-party collaborative learning, for the sake of communication cost or privacy protection, training data from each party cannot be shared with the others. As a result, the dataset for training each local model is confined to its own party and not allowed to be combined with the others to retrain a new model. How to aggregate the knowledge of multiple models into a single model without acquiring their original training datasets is still an open problem. Federated Learning (FL)~\cite{McMahanMRHA17,yang2019federated,chen2019communication,sattler2019robust,9463409,9632275} was recently proposed as a solution to this challenge and has seen remarkable growth. FL introduces a new machine learning paradigm that allows it to learn from distributed data providers without accessing the original data. FL's main workflow is made up of three steps: 1) the server provides the model (\emph{global model}) to the clients; 2) the clients train the model with their own private data and submit the trained model (\emph{local model}) parameters to the server; 3) the server aggregates the local models to produce the new global model. FL's ultimate goal is to develop a global model that works well for all client data after repeating the above three steps for multiple communication rounds. 

However, federated learning requires numerous rounds of communication which costs a lot, thus single-round communication is in high demand in practical applications. In many real-world scenarios, client-side resources are usually limited, the client is unwilling to repeat each round of model updates and instead wants to acquire a model with good performance after only one round of federated learning. As a result, one-shot federated learning gains increasing research interest in the community.


\begin{figure*}[t]
\centering
    \subfloat[]{
    \label{task}
    \includegraphics[width=0.235\linewidth]{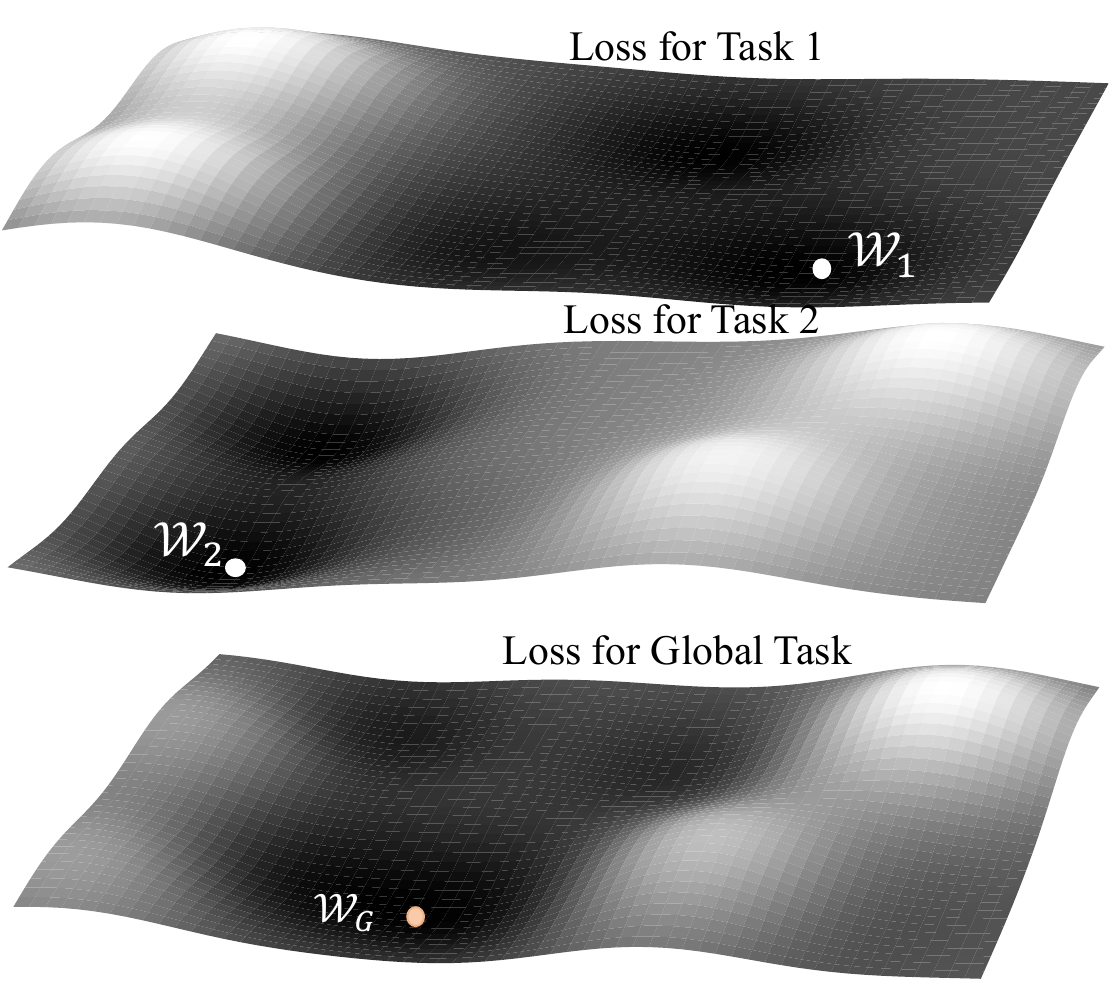}
    }
    \subfloat[]{
    \label{goo}
    \includegraphics[width=0.235\linewidth]{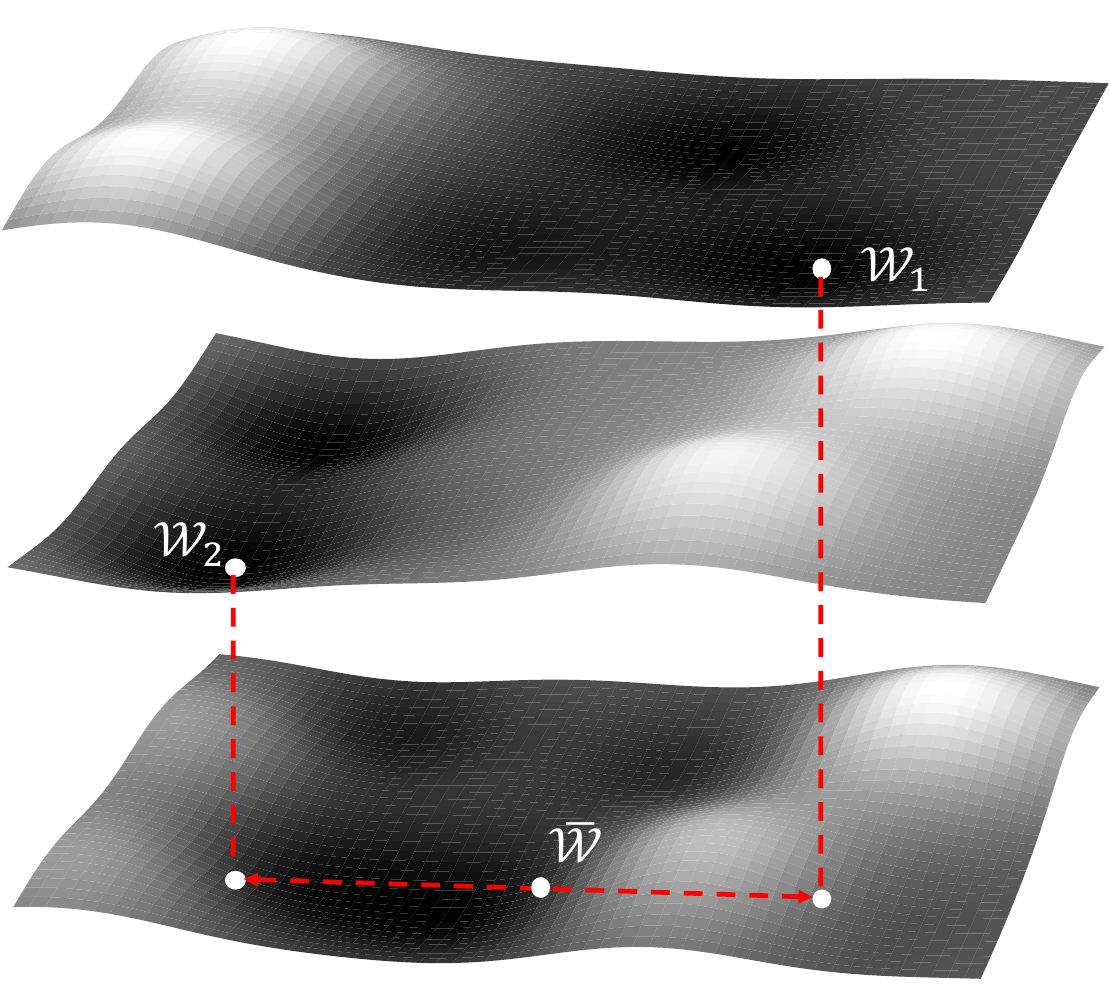}
    }
    \subfloat[]{
    \label{ot}
    \includegraphics[width=0.235\linewidth]{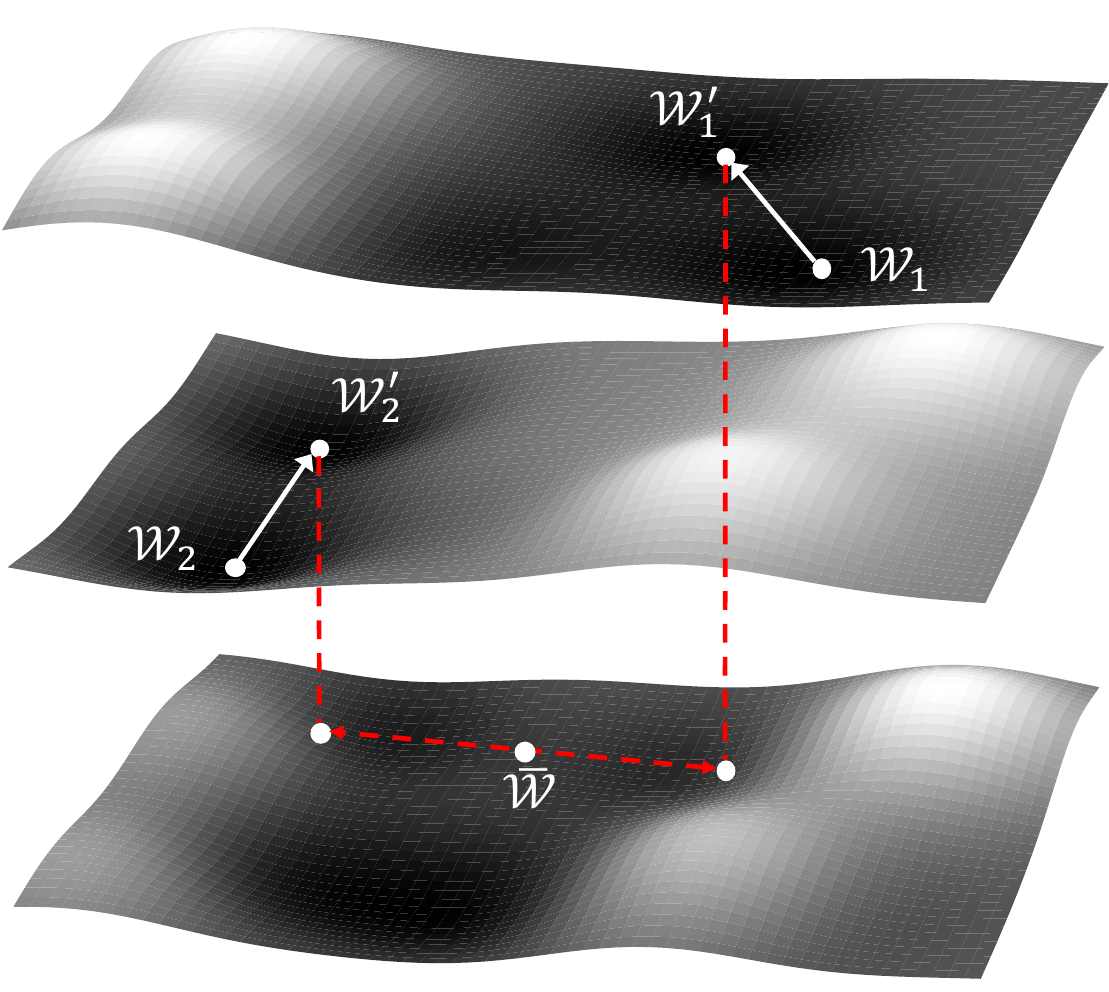}
    }
    \subfloat[]{
    \label{ours}
    \includegraphics[width=0.235\linewidth]{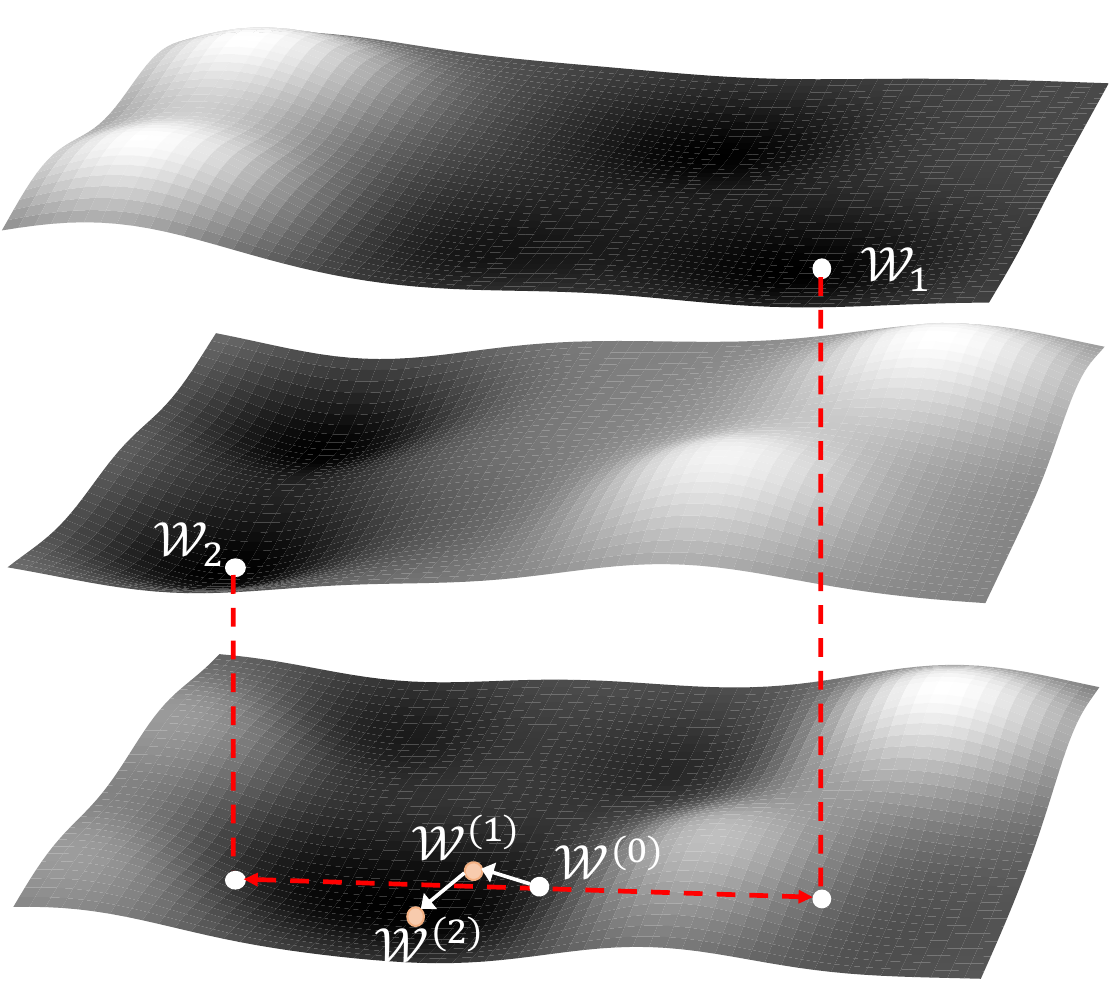}
    }
  \caption{
  We illustrate the model aggregation with simple loss landscapes and different methods to obtain the global model. (a) From top to bottom are the loss surfaces of Task 1, Task 2, and Global Task, respectively. The global task loss value is equal to the sum of Task1 and Task 2. After training Tasks 1 and 2, the local model parameters converge to $\mathcal{W}_1$ and $\mathcal{W}_2$, respectively. Model Aggregation aims to explore a global model $\mathcal{W}_G$, which has low loss for both tasks.  (b) Vanilla average. Directly average without any constraints. Due to the complexity of the loss surface, the falling point $\mathcal{\bar{W}}$ is random. (c) Methods based on neuron-matching. Optimize the permutation matrix, then $\mathcal{W}_i^{'}=\{T^1W_i^1,\cdots,T^LW_i^L\}$. Due to the permutation invariance of the neural network, permutating neurons will not affect the loss value of $\mathcal{W}_1$ and $\mathcal{W}_2$. Finally, an average operation is still needed. (d) Our method directly explores a low-loss global model iteratively based on Algorithm~\ref{algo1}, starting from the initial point $\mathcal{W}^{(0)}$ (can be initialized using Vanilla Average).}
\end{figure*}

Currently, most one-shot federated learning methods~\cite{guha2019one,salehkaleybar2021one,DBLP:conf/ijcai/LiHS21,zhang2021fedzkt,zhou2020distilled,9632275} rely on Knowledge Distillation~\cite{hinton2015distilling} or Dataset Distillation~\cite{wang2018dataset}. The fundamental idea is to use public data sets or the distilled synthetic data from clients to do model distillation on the server, where the local models play as teachers. There are still two issues with this approach. On one hand, it involves an additional training phase when compared to direct averaging, therefore the training cost is expensive. On the other hand, there is a problem of data domain mismatch between clients and the servers. The distillation may cause negative transfer if the public data set and the client data set do not originate from the same domain.

{In order to avoid the aforementioned issues, DENSE~\cite{zhangdense} tried to train a generator on the server side using the client model, so as to generate pseudo samples for multi-teacher distillation. However, the pseudo samples generation stage and distillation stage in DENSE incur a large amount of calculation costs. An alternative approach is to directly extract knowledge from the parameters of local models themselves on the server without model training. In this work, we intend to avoid using public data sets or pseudo samples, as well as incurring additional training costs. }Under this setting, the core issue is how to aggregate local model parameters without training on the server, which is essential to find a \emph{global model} ($\mathcal{W}_G$ in Figure \ref{task}) in the parameter space for all \emph{local models} (which refer to $\mathcal{W}_1$ and $\mathcal{W}_2$ in Figure \ref{task}) such that they can reach a consensus that each of them obtains a relatively low loss on its own dataset. When the original training datasets of local models are available, it is easy to aggregate these models' knowledge by retraining a new model (\emph{ideal global model}) on the union of their datasets. However, it is challenging in the absence of the original training data. 

Without taking into account the public data set and additional training stage, some existing parameter aggregation methods in traditional multi-rounds FL are based on directly averaging \cite{McMahanMRHA17} or cross-model neuron-matching \cite{YurochkinAGGHK19,singh2020model,WangYSPK20}. In multi-rounds FL, the client only trains the local model for a small number of steps, however, in the one-shot FL, the local model must be trained to converge, resulting in significant changes in parameters compared to the model originally received by the client-side. As a result, the direct averaging or neuron-matching strategy cannot be effective. For this reason, considering that the local model has already learned a lot of knowledge, we think from a new perspective, focusing on how to retain as much of the learned knowledge as possible in the procedure of server-side model aggregation.

In this paper, we consider one-shot cross-silo federated learning without server-side training and propose a method for Model Aggregation via Exploring Common Harmonized Optima (MA-Echo). MA-Echo explores the common optima for all local models, and tries to keep the original loss for each local model from being destroyed in aggregation through finding the direction orthogonal to the space spanned by the local data, then the global model can remember what each local model has learned. There is an additional benefit that our method can be naturally combined with the methods based on neuron-matching. We conduct extensive experiments on two popular image classification datasets, MNIST and CIFAR-10, to validate the effectiveness of the proposed MA-Echo. We also visualize the changing trajectory of the global model $\mathcal{W}_G$ during the iteration procedure and empirically test the robustness of MA-Echo on the training datasets with varying degrees of non-identicalness. In order to visually observe whether the knowledge in the model is retained, we aggregate multiple conditional variational auto-encoders (CVAE)~\cite{NIPS2015_8d55a249} and check whether the aggregate model can generate images in categories that the original single local model has never learned. According to the results, MA-Echo outperforms the other methods and works well even in extremely non-identical data distribution settings; meanwhile, the application of our aggregation method can empirically improve the convergence speed of multi-rounds federated learning. Our contributions can be summarized as follows:
	\begin{itemize}
		\item We consider a challenging but realistic one-shot federated learning problem setting where model training on a {public or pseudo dataset} is not allowed on the server. 
		\item Under this setting, we propose an effective method MA-Echo, which preserves the knowledge of the local model parameters in the global model as much as possible during the model aggregation procedure on the server. 
		\item We conduct extensive experiments to evaluate the performance of MA-Echo compared with parameters averaging and neuron-matching based aggregation methods, and find that it significantly outperforms the other state-of-the-art aggregation algorithms.
	\end{itemize}

\section{Related Work}
\textbf{Federated Learning.} FL is a new machine learning setting under privacy protection, first proposed in~\cite{konevcny2016federated} and~\cite{McMahanMRHA17}. FedAvg~\cite{McMahanMRHA17} is the most basic and widely used FL method. There are several attempts to improve FedAvg to adapt to the Non-IID local datasets. As one of the earliest variants of FedAvg, FedProx~\cite{conf/mlsys/LiSZSTS20} introduces a dynamic regularization term in the client loss, so that each local model would be closer to the global model. Recent work FedDyn~\cite{acar2021federated} also proposes a dynamic regularization method to handle the problem that the minima of the local empirical loss are inconsistent with those of the global empirical loss. {FedBN~\cite{li2021fedbn} discards the aggregation of BN layers when doing federated averaging, and achieves better performance in cross domain datasets.} There are some other works that have proposed improved methods for Non-IID settings. SCAFFOLD~\cite{karimireddy2020scaffold} uses control variates to reduce client drift. From the optimization perspective,~\cite{reddi2020adaptive} applies momentum on the server aggregation stage and proposes FL versions of AdaGrad, Adam, and Yogi. Based on the Non-IID setting, FedNova~\cite{wang2020tackling} aims to solve the heterogeneous local progress problem, where each client has heterogeneous local updates, by normalizing the local models according to the local updates before averaging. {FedGen~\cite{zhu2021data} uses the local models and local category distributions uploaded from the clients to train a generator, $G(\cdot)$. Then the server sends $G$ to the clients, where additional data features $z \sim G(\cdot |y)$ are generated for improving local training.} It is worth mentioning that, recently FedMIX~\cite{yoonfedmix} proposes to share the averaged local data among clients; and discusses the privacy risks from the averaged local data. In our work, we will collect the projection matrix from each client, which can be understood as a kind of auxiliary information of the local data. 


Some works~\cite{mohri2019agnostic,li2019fair,hu2020fedmgda+} propose to study fair aggregation techniques. They aim to optimize the weighting coefficients to make the aggregation step be with more fairness. FedMGDA+~\cite{hu2020fedmgda+} also handles FL from the multi-objective optimization perspective, but it is different from the proposed MA-Echo because we construct a new optimization objective according to our motivation instead of the straightforward optimization goal used in~\cite{hu2020fedmgda+}. In addition, there are some methods that do not use model averaging. FedDF~\cite{lin2020ensemble} combines FL with Knowledge Distillation~\cite{hinton2015distilling} and trains the global model with some extra data or pseudo samples on the server. MOON~\cite{li2021model} uses Contrastive Learning~\cite{chen2020simple} in the clients to improve the local training of each parties. FedMA~\cite{WangYSPK20} and OT~\cite{singh2020model} perform neuron matching to improve the global model. 

\textbf{One-shot Federated Learning.} There are also some works trying to study one-shot FL. \cite{guha2019one,DBLP:conf/ijcai/LiHS21} use Knowledge Distillation~\cite{gou2021knowledge} technology. They use the local models as teachers, and use public data sets or pseudo samples generated from the local models to train a student model. {DENSE~\cite{zhangdense} divides the aggregation process on the server side into two steps. The first step is to train a generator using the client models. The second step is  multi-teacher knowledge distillation using the pseudo samples generated by the generator. } \cite{zhou2020distilled} use Dataset Distillation~\cite{wang2018dataset} in clients and upload the distilled synthetic data to the server to train the global model. \cite{shin2020xor} propose to collect the encoded data samples from clients and then use samples on the server to decode the collected data, the decoded samples are used to train the global model. \cite{salehkaleybar2021one} gives some theoretical analysis of one-shot FL, but limited to the i.i.d. case. It is worth mentioning that the methods introduced above are not in conflict with our method, and their results can be used as the initial iteration point for MA-Echo.

\textbf{Model Aggregation.} There are some aggregation methods in traditional multi-rounds FL that do not require public data sets and additional training processes. Google proposes FedAvg~\cite{McMahanMRHA17} to directly average the parameters of local models to obtain the global model. \cite{YurochkinAGGHK19} notices the permutation invariance of neurons across models and proposes PFNM, which employs a Bayesian nonparametric approach to aligning the neurons between multilayer perceptrons layer by layer. \cite{singh2020model} uses the Optimal Transport (OT) algorithm to match neurons across models in each layer directly and then averages the re-aligned local model parameters. \cite{WangYSPK20} further improves the PFNM by proposing the FedMA and applies it to more complex architectures such as CNNs. When different local models have the same model structure,  which is the default setting for model aggregation, FedMA is equivalent to finding a Wasserstein barycenter~\cite{journals/siamma/AguehC11} in all local models, which is similar to~\cite{singh2020model}. These methods based on cross-model neuron-matching all permute the rows of the parameter matrix in each layer of a neural network to achieve a good match (i.e., to reduce the distance between local models, see Section~\ref{pre}) and then aggregate the local models via direct averaging.

{
\section{Problem Setting}}

Suppose we have $N$ models $\{f_i\}_{i=1}^N$ that are trained on $N$ client-datasets $\{\mathcal{D}_i\}_{i=1}^N$ to convergence, respectively. We call them local models. The parameters of the $i$-th local model are composed of $L_i$ layers, denoted by $\mathcal{W}_i = \{ {W_i^1,\dots,W_i^{L_i}}\}$, where $W_i^l \in \mathbb{R}^{C^l_{out} \times C^l_{in}}$, $C^l_{out}$ is the number of output dimensions, the $h$-th row of $W_i^l$ represents the parameter vector corresponding to the $h$-th output neuron. In this paper, we consider that these local models have the same architecture, which means $L_1=L_2=\ldots=L_N$ and that, for each layer $l$, $\{W_i^l\}_{i=1}^N$ have the same size. The aim of our setting is to return the global model parameters $\mathcal{W}_G = \{ {W_G^1,\dots,W_G^{L}}\}$ which should have the classification ability of any local model without acquiring the local training datasets or some public dataset, and there is no training phase on the server.

{
\section{Preliminaries}
\label{pre}}

{
In the following, we first briefly introduce two baseline aggregation methods, vanilla average and neuron-matching based methods. Then we will introduce the concept of null space projection, which will be employed in our technique.
}
\textbf{Vanilla average.} For the model aggregation, a straightforward method is to average the local model parameters directly (see Figure\ref{goo}). However, there is no guarantee to make the averaged model close to the low-loss area.
 
\textbf{Neuron-matching based methods.} These methods ~\cite{singh2020model,WangYSPK20} observe that changing the permutation of neurons (i.e., dimensions of a model parameter vector) does not affect model performance. The core idea is to solve this optimization problem:
\begin{linenomath}
\begin{align}
 &\min _{W^l} \sum_{i=1}^{N} \min _{T_{i}^{l}} R\left(W^l, T_{i}^{l} W_{i}^{l}\right)
\label{match}
\end{align}
\end{linenomath}
where $T_{i}^{l}$ is a permutation matrix (there is only one `1' in each row and each column while the rest of entries are `0'), $R(A,B)= ||A-B||^2_F$. Then the optimal matrix $T_i^{l*}$ is used to calculate the global model: $W_G^l=\frac{1}{N}\sum_i T_{i}^{l*} W_{i}^{l}$. 

We can understand the goal of this type of method like this: Align neurons between local models through permutation to make them as close as possible. According to the local smoothness assumption, as long as these local models are close enough (in the same low-loss area, i.e. the same valley of the loss landscape), one can obtain a better global model after the parameters are averaged. However, on one hand, the optimization problem on shortening the distance between two models is largely restricted by the specific structure of the permutation matrix (only one `1' in each row and column). There is always a fixed shortest distance between two models in this particular discrete optimization problem. On the other hand, the update of the local models in one-shot FL is large, resulting in a large distance between the local models. Once the local models have not been close enough after optimization mentioned above, the performance of the global model cannot be guaranteed through averaging (because the average is more likely to fall in a higher-loss point if the convex hull constituted by the local models is not small enough).

\textbf{Null space projection.}
Consider a simple parameter vector $\mathbf{w}\in \mathbb{R}^{d}$, the input for $\mathbf{w}$ is $X\in \mathbb{R}^{n \times d }$, the output is $\mathbf{y}=X\mathbf{w}$. When we impose a disturbance quantity $\Delta\mathbf{w}$ on $\mathbf{w}$, we have $X(\mathbf{w}+\Delta\mathbf{w})=\mathbf{y}+X\Delta\mathbf{w}$. In order to keep the original mapping unchanged, $X\Delta\mathbf{w}$ should be equal to $\mathbf{0}$, which means $\Delta\mathbf{w}$ should lie in the null space of $X$. Therefore, we need to project $\Delta\mathbf{w}$ into the null space of the input feature. Due to the linear transformation, the projection has a fixed form: $\Delta\mathbf{w}\leftarrow (I-P)\Delta\mathbf{w}$, where $P = X^\top(XX^\top+zI)^{-1}X$, $z$ is a small constant for avoiding the ill-conditioning issue in the matrix-inverse operation. The projection matrix has been applied into Continual Learning~\cite{zeng2019continual,farajtabar2020orthogonal,ChaudhryOrthogSubspaceCL}. In order to reduce the computational complexity, we use iterative method~\cite{zeng2019continual} to calculate $P$.

\section{MA-Echo}
\label{method}

\subsection{The Proposed Objective}
\label{method_1}

 For brevity of description, we start with a simple parameter vector $\mathbf{w}$, which can be easily generalized to matrix form. The most straightforward method for model aggregation is to minimize the objective function of these local models simultaneously from the multi-objective optimization perspective:
\begin{linenomath}
\begin{align}
   \label{mom}
    \mathbf{w}_G = \arg\min_{\mathbf{w}} \mathcal{L} \triangleq \left[\mathcal{L}_{1}(\mathbf{w}), \mathcal{L}_{2}(\mathbf{w}), \ldots \mathcal{L}_{N}(\mathbf{w})\right]^{\top}
\end{align}
\end{linenomath}
However, due to the inaccessibility of the training data, we cannot directly optimize these objective functions in the aggregation step. Moreover, under the one-shot setting, the client can only perform one complete training, so FedAvg's multi-round distributed iterative optimization cannot be used. In this paper, we consider model aggregation as a forgetting-alleviation problem. Suppose the $i$-th local model parameter is $\mathbf{w}_i$, then we want to find the global parameter $\mathbf{w}_G$, which can remember what each local model has learned. To alleviate the forgetting problem of $\mathbf{w}_G$, we let each $\mathbf{w}_G-\mathbf{w}_i$ be in the direction orthogonal to the space spanned by the input feature:
\begin{linenomath}
\begin{align}
   \label{mom2}
    \min _{\mathbf{w}}\left[\left\|P_{1}\left(\mathbf{w}-\mathbf{w}_{1}\right)\right\|_{2}^{2}, \ldots, \left\|P_{N}\left(\mathbf{w}-\mathbf{w}_{N}\right)\right\|_{2}^{2}\right]^{\top}
\end{align}
\end{linenomath}
where $P_{i}$ is the projection matrix of the $i$-th local model. When $\left\|P_{i}\left(\mathbf{w}-\mathbf{w}_{i}\right)\right\|_{2}$ approachs to $0$, $\left(\mathbf{w}-\mathbf{w}_{i}\right)$ will be orthogonal to the space spanned by the input feature of the $i$-th local model, as shown in Preliminaries, $\mathbf{w}$ will not forget the knowledge of $\mathcal{D}_i$ learned by $\mathbf{w}_i$. With the help of the projection matrix, we can liberate the aggregation from multiple rounds of communication and only do it on the server. 

Note that in the neural network, there are often a large number of local optimal solutions, so for the neural network, $\mathbf{w}_i$ in Eq.\ref{mom2} can be any local optimal solution. We define $\mathcal{S}_{\mathbf{w}_i}$ as a local optimal solution set, where each element has a loss value similar to $\mathbf{w}_i$. Then Eq.\ref{mom2} is rewritten as:
\begin{linenomath}
\begin{align}
   \label{mom3}
    \nonumber\min_{\mathbf{w},\{\mathbf{v}_i\}}&\left[\left\|P_{1}\left(\mathbf{w}-\mathbf{v}_{1}\right)\right\|_{2}^{2}, \ldots, \left\|P_{N}\left(\mathbf{w}-\mathbf{v}_{N}\right)\right\|_{2}^{2}\right]^{\top}\\
    &\text { s.t. } \mathbf{v}_i \in \mathcal{S}_{\mathbf{w}_i},  i = 1,\cdots,N
\end{align}
\end{linenomath}

\subsection{The Proposed Solution }
We alternately optimize Eq.\ref{mom3}. First, we fixed $\mathbf{v}_i$ and optimize $\mathbf{w}$. We use the gradient-based method to solve the problem: $\mathbf{w}^{(t+1)}= \mathbf{w}^{(t)}+\eta \mathbf{d}^{(t)}$. Following \cite{fliege2000steepest} which introduces a simple method to find $\mathbf{d}$, we expect that $\mathbf{d}^{(t)}$ can reduce each sub-objective in Eq.\ref{mom3}, that is, the inner product of $-\mathbf{d}^{(t)}$ and the gradient of each sub-objective $ P_i^\top P_i(\mathbf{w}-\mathbf{v}_i)$ should be as large as possible (i.e. minimize $\left(\mathbf{w}-\mathbf{v}_{i}\right)^{\top}  P_{i}^{\top} P_{i} \mathbf{d}$). Then we introduce the slack variable $\epsilon$, so that multiple objectives can be adjusted adaptively. Finally we formulate the descent direction as a constrained optimization problem:
\begin{linenomath}
\begin{align}
\label{mind}
&\min _{\mathbf{d}, v, \epsilon_{i}} v+\frac{1}{2}\|\mathbf{d}\|_{2}^{2}+C \sum_{i=1}^{N} \epsilon_{i} \\\nonumber
\text { s.t. }2\left(\mathbf{w}-\mathbf{v}_{i}\right)^{\top} & P_{i}^{\top} P_{i} \mathbf{d} \leq v+\epsilon_{i},  \epsilon_{i} \geq 0, i = 1,\cdots,N
\end{align}
\end{linenomath}
where $v$ is used to minimize the upper bound of all inner products, the slack variable $\epsilon$ relax global models to forget some knowledge during the aggregation process adaptively. By Lagrange multipliers, the dual problem of Eq.\ref{mind} is:
\begin{linenomath}
\begin{align}
\label{svm}
    &\underset{\boldsymbol{\alpha}}{\operatorname{min}} \frac{1}{2}\left\|\sum_{i=1}^{N} 2 \alpha_{i} P_{i}^{\top} P_{i}\left(\mathbf{w}-\mathbf{v}_{i}\right)\right\|_{2}^{2} \\ \nonumber \text { s.t. } &\sum_{i=1}^{N} \alpha_{i}=1,0\leq\alpha_{i}\leq C, i = 1,\cdots,N
\end{align}
\end{linenomath}
where $\alpha_i$ denotes the Lagrange multiplier of the first inequality constraint in Eq.\ref{mind}, {as $P$ is a projection matrix, using its definition in Section~\ref{pre}, we have $P_{i}^{\top} P_{i}=P_{i}$ }. It is interesting that Eq.\ref{svm} is a One-class SVM problem, we can thus use any open source library  to solve it (we use CVXOPT~\cite{andersen2013cvxopt} in this paper). Finally we have:
\begin{linenomath}
\begin{align}
\label{grad}
\mathbf{d}^{(t)}=-\sum_{i=1}^{N} 2 \alpha_{i}^{*} P_{i}\left(\mathbf{w}^{(t)}-\mathbf{v}_{i}\right)
\end{align}
\end{linenomath}
where $\boldsymbol{\alpha}^*$ is the solution of Eq.\ref{svm}. Repeat Eq.\ref{grad} and $\mathbf{w}^{(t+1)}= \mathbf{w}^{(t)}+\eta \mathbf{d}^{(t)}$ several times, we can find the approximate solution of $\mathbf{w}$.

Second, we fix $\mathbf{w}$ and optimize $\mathbf{v}_i$ in Eq.\ref{mom3}. Since the parameters of each local model are independent, we can optimize each sub-objective separately:
\begin{linenomath}
\begin{align}
\label{optv}
    \min _{\mathbf{v}_{i}}\left\|\left(\mathbf{w}-\mathbf{v}_{i}\right)\right\|_{2}^{2}+\mu\left\|P_i\left(\mathbf{v}_{i}-\mathbf{w}_{i}\right)\right\|_{2}^{2}
\end{align}
\end{linenomath}
{Note that in the first term, if we retain a projection matrix, i.e., using $\left\|P_i\left(\mathbf{w}-\mathbf{v}_i\right)\right\|_2^2$, it will be difficult to obtain an analytical solution. However, for any orthogonal projection matrix $P$, we have $\left\|P\left(\mathbf{w}-\mathbf{v}_i\right)\right\|_2^2< \left\|\left(\mathbf{w}-\mathbf{v}_i\right)\right\|_2^2$, then we can indirectly optimize the upper bound $\left\|\left(\mathbf{w}-\mathbf{v}_i\right)\right\|_2^2$. The second term is to ensure that $\mathbf{v}_i$ is in the solution set $ \mathcal{S}_{\mathbf{w}_i}$, when the second term approaches to $0$, $\mathbf{v}_i$ has the same loss value as $\mathbf{w}_i$.} Let the derivative of the objective be $0$, we get:
\begin{linenomath}
\begin{align}\nonumber
    \mu P_i^{\top} P_i\left(\mathbf{v}_{i}-\mathbf{w}_{i}\right)+\mathbf{v}_{i}-\mathbf{w}=0
\end{align}
\end{linenomath}
Note that $P_i$ is a projection matrix, so $P_i^\top P_i=P_i$ and $P_i^\top=P_i$, we have $\mu P_i\left(\mathbf{v}_{i}-\mathbf{w}_{i}\right)+\mathbf{v}_{i}-\mathbf{w}=0$, then:
\begin{linenomath}
\begin{align}
\label{inv}
\mathbf{w}-\mathbf{w}_{i}=\mu P_{i}\left(\mathbf{v}_{i}-\mathbf{w}_{i}\right)+\mathbf{v}_{i}-\mathbf{w}_{i}=\left(I+\mu P_{i}\right)\left(\mathbf{v}_{i}-\mathbf{w}_{i}\right)
\end{align}
\end{linenomath}
Note that $P_i$ satisfies $P_i^2=P_i$, and if $\mu<1$ then $(\mu P_i)^n\to 0$, by Taylor expansion, we have:
\begin{linenomath}
\begin{align}
\label{tay}
\left(I+\mu P_{i}\right)^{-1}=I-\mu P_{i}+\mu^{2} P_{i}-\mu^{3} P_{i} \ldots \approx I-\frac{\mu P_{i}}{1+\mu}
\end{align}
\end{linenomath}
take Eq.\ref{tay} into Eq.\ref{inv}, we have:
\begin{linenomath}
\begin{align}
\label{sec}
    \mathbf{v}_{i} =\mathbf{w}_i+ \left(I-\frac{\mu P_{i}}{1+\mu}\right)\left( \mathbf{w}-\mathbf{w}_i\right)
\end{align}
\end{linenomath}
where we take $\mu=1$ by default.

\begin{algorithm}[t]
\caption{MA-Echo}\label{algo1}
\begin{algorithmic}
\STATE {{\bfseries Input:} $\{W_i^1,\dots,W_i^L\}_{i=1}^{N}$, $\{P_i^1,\dots,P_i^L\}_{i=1}^{N}$, $\tau$, $\eta$.}\\
\STATE {{\bfseries Output:} Global model parameters $\{W_G^1,\dots,W_G^L\}$ }

\STATE $t=0$, $W^{l(0)}=\frac{1}{N} \sum_{i=1}^N W^{l}_i$, $V^l_i=W^l_i$ for $l=1,\cdots,L$.
\WHILE{$t < \tau$}
    \FOR{$l$ from 1 to $L$}
    	\STATE $\boldsymbol{\alpha}^*\leftarrow\text{Solve}\,\, Eq.\ref{svm}.$ \text{in matrix form}\\
		\STATE $D^{l(t)}=-\sum_{i=1}^{N} 2 \alpha_{i}^{*}  \left(W^{l(t)}-V^l_{i}\right)P^l_{i} $\\
		\STATE $W^{l(t+1)}=W^{l(t)}+\eta D^{(t)}$
        \FOR{$i$ from 1 to $N$}
            \STATE $V^l_i \leftarrow V^l_i+\text{Norm}((W^{l(t+1)}-V^l_i)(I-\frac{1}{2}P^l_i))$
        \ENDFOR
    \ENDFOR
\ENDWHILE
\RETURN $\{W^{1(\tau)},\dots,W^{L(\tau)}\}$
\end{algorithmic}
\end{algorithm}

We deal with the neural network with a layer-wise treatment. For multilayer perceptron, the $l$-th layer is $W_i^l$, the derivation process mentioned above still holds, then the matrix form results (Eq.\ref{grad} and Eq.\ref{sec}) can be rewritten as:
\begin{linenomath}
\begin{align}
\nonumber
D^{l(t)}&=-\sum_{i=1}^{N} 2 \alpha_{i}^{*}  \left(W^{l(t)}-V^l_{i}\right)P^l_{i}\\
\nonumber
V^l_i &= W^l_i+(W^l-W^l_i)(I-{\frac{\mu}{1+\mu}}P^l_i)
\end{align}
\end{linenomath}
The overall method is in Algorithm~\ref{algo1}. In the first step, we use the average parameters of the local models as the initial point of optimization iterations in MA-Echo. Also, we have a variety of initialization methods that can be used. In Section~\ref{experiments}, we show three different initialization strategies. 

For convolutional neural networks, $W_i^l \in \mathbb{R}^{C_{out} \times C_{in}\times h\times w}$ ($h$ and $w$ are the length and width of the convolution kernel, respectively; $C_{out}$ and $C_{in}$ are the number of output channels and input channels of the convolution layer, respectively), we can reshape it to $\hat{W}_i^l \in \mathbb{R}^{C_{out} \times (C_{in}*h*w)}$, then subsequent calculations are the same as the fully-connected layer in the multilayer perceptron. We provide an optional operation $\text{Norm}(\cdot)$, where $\text{Norm}(
W)=W$ or $\text{Norm}(
W)= \text{torch.norm}(W,\text{dim}=1)$, we find that the normalized parameter update can make the algorithm more stable.

\subsection{Applications}
\textbf{Work together with neuron-matching based methods} Suppose we have the optimal permutation matrix $T^*$ of Eq.\ref{match} in the $(l-1)$-th layer, due to the parameter permutation, the input vectors of the $l$-th layer $X$ are changed to $X^{'}=XT^*$. For that $T^*{T^*}^\top =I$, we have:
\begin{linenomath}
\begin{align*}
    P^{'} =\,\,{X^{'}}^\top(X^{'}{X^{'}}^\top)^{-1}X^{'}
    =\,\,{T^*}^\top X(X {T^*} {T^*}^\top X^\top)^{-1}X{T^*}={T^*}^\top P {T^*} 
\end{align*}
\end{linenomath}
So our method does not conflict with the neuron-matching based method. When we get ${T^*}$, then $W$ and $P$ can be updated by $W \leftarrow {T^*}W$ and $P \leftarrow {T^*}^\top P {T^*} $, the subsequent steps are the same as Algorithm \ref{algo1}.

\textbf{Applied to Multi-round Federated Learning.} A large number of FL algorithms~\cite{McMahanMRHA17,conf/mlsys/LiSZSTS20,karimireddy2020scaffold,wang2020tackling} study how to accelerate the overall optimization speed of multi-round communication, but very few works try to develop more effective parameter aggregation algorithms within a single-round communication. The MA-Echo in this paper can be directly used to replace the parameter averaging operation in federated learning. In each communication round, the server sends the global model (the output of Algorithm~\ref{algo1}) to each client, then the clients retrain the model based on their own datasets. We will verify the effect of MA-Echo under the multi-round federated learning setting in our experiments. 

\subsection{Theoretical Analysis}
We do some analysis on MA-Echo in this subsection. For brevity of description,  we still use the vector form $\mathbf{w}$.
\begin{proposition}[Properties of Eq.\ref{grad}]
\label{theo1}
Given the solution of Eq.\ref{mind} being $\left(\mathbf{d}^{*}, v^{*}, \epsilon^{*}\right)$:
\begin{enumerate}
  \item If $\mathbf{w}$ is Pareto critical, then $\mathbf{d}=0$;
  \item If $\mathbf{w}$ is not Pareto critical, then for $i=1, \ldots, N$,
    \begin{eqnarray}\label{the1}
    \begin{linenomath}
    \begin{aligned}
\left\langle P_{i}^{\top} P_{i}\left(\mathbf{w}-\mathbf{v}_{i}\right), \mathbf{d}^{*}\right\rangle \leq v^{*}+\epsilon^{*}=-\left\|\mathbf{d}^{*}\right\|_{2}^{2}+\epsilon_{i}^{*}-C \sum_{n=1}^{N} \epsilon_{n}^{*}
    \end{aligned}
    \end{linenomath}
    \end{eqnarray}
\end{enumerate}
\end{proposition}

\begin{proof}
The Lagrange function of Eq.\ref{mind} is:
\begin{linenomath}
\begin{align}
\label{lag}
    \nonumber\mathcal{L}=&v+\frac{1}{2}\|\mathbf{d}\|_{2}^{2}+C \sum_{i=1}^{N} \epsilon_{i}+\sum_{i=1}^{N} 2 \alpha_{i}\left(\mathbf{w}-\mathbf{v}_{i}\right)^{\top} P_{i}^{\top} P_{i} \mathbf{d}\\&
    -\sum_{i=1}^{N} \alpha_{i} v-\sum_{i=1}^{N} \alpha_{i} \epsilon_{i}-\sum_{i=1}^{N} \gamma_{i} \epsilon_{i}
\end{align}
\end{linenomath}
where $\alpha_i$ and $\gamma_i$ are Lagrange multipliers. Calculating the partial derivative of $\mathcal{L}$ with respect to $\mathbf{d}$, $v$ and $\epsilon$:
\begin{linenomath}
\begin{align}
\nonumber\frac{\partial \mathcal{L}}{\partial \mathbf{d}}=\mathbf{d}+\sum_{i=1}^{N} 2 \alpha_{i} P_{i}^{\top} P_{i}\left(\mathbf{w}-\mathbf{v}_{i}\right) &\\\nonumber
\frac{\partial \mathcal{L}}{\partial v}=1-\sum_{i=1}^{N} \alpha_{i},\quad\frac{\partial \mathcal{L}}{\partial \epsilon_{i}}=C-\alpha_{i}-\gamma_{i} &
\end{align}
\end{linenomath}
Let the derivative equal $0$, we have:
\begin{linenomath}
\begin{align}
\label{soul}
\mathbf{d}=-\sum_{i=1}^{N} 2 \alpha_{i} P_{i}^{\top} P_{i}\left(\mathbf{w}-\mathbf{v}_{i}\right),
\sum_{i=1}^{N} \alpha_{i}=1,
C=\alpha_{i}+\gamma_{i}
\end{align}
\end{linenomath}
Bring Eq.\ref{soul} into Eq.\ref{lag}, then we have the dual problem:
\begin{linenomath}
\begin{align}
\label{dual}
    \underset{\boldsymbol{\alpha}}{\operatorname{min}} \frac{1}{2}\left\|\sum_{i=1}^{N} 2 \alpha_{i} P_{i}^{\top} P_{i}\left(\mathbf{w}-\mathbf{v}_{i}\right)\right\|_{2}^{2}, \nonumber\\  \text { s.t. } \sum_{i=1}^{N} \alpha_{i}=1,0\leq\alpha_{i}\leq C, i = 1,\cdots,N
\end{align}
\end{linenomath}
Denote the solution of Eq.\ref{mind} and Eq.\ref{dual} as $\left(\mathbf{d}^{*}, v^{*}, \epsilon_i^{*}\right)$ and $\left( \alpha_i^{*}, \gamma_i^{*}\right)$. According to the KKT condition and Eq.\ref{mind}, we have:
\begin{linenomath}
\begin{align}
\label{kkt}
    \alpha_{i}^{*}\left(2\left(\mathbf{w}-\mathbf{v}_{i}\right)^{\top} P_{i}^{\top} P_{i} \mathbf{d}^{*}-v^{*}-\epsilon_{i}^{*}\right)=0,  \gamma_{i}^{*} \epsilon_{i}^{*}=0
\end{align}
\end{linenomath}
For that $C=\alpha_{i}+\gamma_{i}$ (Eq.\ref{soul}) and $\gamma_{i}^{*} \epsilon_{i}^{*}=0$, we also have $\alpha_{i}^{*} \epsilon_{i}^{*}=C\epsilon_{i}^{*}$. If $\mathbf{d}^*=0$, obviously all $\left\langle P_{i}^{\top} P_{i}\left(\mathbf{w}-\mathbf{v}_{i}\right), \mathbf{d}^{*}\right\rangle =0$, then $\mathbf{d}^*=0$ corresponds to the Pareto critical point. If $\mathbf{d}^*\neq0$, from Eq.\ref{kkt}, we have:
\begin{linenomath}
\begin{align}
    &\nonumber\sum_{i=1}^m \alpha_{i}^{*}\left(2\left(\mathbf{w}-\mathbf{v}_{i}\right)^{\top} P_{i}^{\top} P_{i} \mathbf{d}^{*}-v^{*}-\epsilon_{i}^{*}\right)\\\nonumber&=\sum_{i=1}^{N} \alpha_{i}^{*} 2\left(\mathbf{w}-\mathbf{v}_{i}\right)^{\top} P_{i}^{\top} P_{i} \mathbf{d}^{*}-\sum_{i=1}^{N} \alpha_{i}^{*} v^{*}-\sum_{i=1}^{N} \alpha_{i}^{*} \epsilon_{i}^{*}=0
\end{align}
\end{linenomath}
Bring Eq.\ref{soul} into the above equation, we have:
\begin{linenomath}
\begin{align}
\label{v8}
    -\left\|\mathbf{d}^{*}\right\|_{2}^{2}-v^{*}-\sum_{i=1}^{N} C \epsilon_{i}^{*}=0
\end{align}
\end{linenomath}
take Eq.\ref{v8} into the constraint of Eq.\ref{mind}, we can complete this proof:
\begin{linenomath}
\begin{align}\nonumber
2\left(\mathbf{w}-\mathbf{v}_{i}\right)^{\top} P_{i}^{\top} P_{i} \mathbf{d}^{*} \leq v^{*}+\epsilon_{i}^{*}=-\left\|\mathbf{d}^{*}\right\|_{2}^{2}+\epsilon_{i}^{*}-\sum_{i=1}^{N} C \epsilon_{i}^{*}
\end{align}
\end{linenomath}
\end{proof}

Proposition~\ref{theo1} reveals the properties that the solution of Eq.\ref{grad} satisfies. From Eq.\ref{svm}, we know that $C\in [1/N,1]$ and then from Eq.\ref{the1}, we can see: 1) when $C=1$, for each $i$, $\left\langle P_{i}^{\top} P_{i}\left(\mathbf{w}-\mathbf{v}_{i}\right), \mathbf{d}^{*}\right\rangle < 0$, so that each sub-objective of Eq.\ref{mom2} gets decreased, which means that $\mathbf{w}$ is trying to remember the knowledge of all local models. 2) when $C=1/N$, then $\alpha_i=1/N$, each local model is equally important. 3) when $C\in (1/N, 1)$, the $\alpha_i^*$ will vary for different local models, so that $\mathbf{w}$ can adaptively forget part of the knowledge of some local models. 

When applying MA-Echo to the multiple-round federated learning setting, we explore the convergence properties of the algorithm in a relatively simple situation, where local training has $1$ epoch and full batchsize. Let the global model in the $k$-th communication round be $\mathbf{w}_{(k)}$, then the server sends $\mathbf{w}_{(k)}$ to the clients. Each client trains the local model and send the update $g^i_k$ to the server, then we get $\mathbf{w}_{(k)}-\lambda_{k}\hat{g}_{k}$ on the server, where $\hat{g}_{k}=1/N\sum^N_{i=1} g^i_k$. Run MA-Echo, we have $\mathbf{w}_{(k+1)}=\mathbf{w}_{(k)}-\lambda_{k} \hat{g}_{k}+\eta_{k}\hat{d}_{k}$, where $\hat{d}_{k}$ is derived from the iterations in the aggregation. After multiple rounds of communication, the distance between $\mathbf{w}_{(k)}$ and the optimal solution is as follows:
\begin{proposition}
Suppose that each local model is M-Lipschitz continuous and $\sigma$-strongly convex, and $\hat{d}_{k}$ is upper bounded: $\|\hat{d}_{k}\|^{2} \leq G^{2}$. With $Epoch=1, Batchsize=|\mathcal{D}_i|$ for local training  and the choices of $\lambda_{k}=\frac{3}{c(k+2)}$, $\eta_{k}=\frac{1}{(k+2)(k+1)}$, we have:
\begin{linenomath}
\begin{align}
\label{the2}
   \mathbb{E}\left[\left\|\mathbf{w}_{(k+1)}-\mathbf{w}_{(k+1)}^{*}\right\|^{2}\right] \leq
    \left[\frac{2}{(k+1)(k+2)}+\frac{1}{2(k+2)}\right] G^{2}+
    \frac{18 M^{2}}{c^{2}(k+2)}
\end{align}
\end{linenomath}
where $c$ is a constant, $\mathbf{w}_{(k+1)}^{*}$ is the projection of $\mathbf{w}_{(k+1)}$ to the Pareto stationary set $\mathcal{M}$, i.e., $\mathbf{w}_{(k+1)}^{*}=\underset{\mathbf{w} \in \mathcal{M}}{\operatorname{argmin}}\|\mathbf{w}_{(k+1)}-\mathbf{w}\|$.
\end{proposition}

\begin{proof}
Use $I_i\in \{0,1\}^N$ to denote which client we sample in each round, we define $\hat{L}_i(\mathbf{w},I)=I_i L_i(\mathbf{w})$,
then the objective of FL is equivalent to:
\begin{linenomath}
\begin{align}
\label{mom_m}
    \min _{\mathbf{w}}\left\{\hat{L}_{i}(\mathbf{w}, I), \ldots, \hat{L}_{N}(\mathbf{w}, I)\right\}
\end{align}
\end{linenomath}
suppose $\mathbf{w}^*_{(k)}$ is the projection of $\mathbf{w}_{(k)}$ to the Pareto stationary set of Eq.\ref{mom_m}, then $\mathbf{w}_{(k+1)}^{*}=\underset{\mathbf{w} \in \mathcal{W}}{\operatorname{argmin}}\|\mathbf{w}_{(k+1)}-\mathbf{w}\|$, so that:
\begin{linenomath}
\begin{align}
\label{abs}
    \forall i, \hat{L}_{i}\left(\mathbf{w}_{(k)}, I_{k}\right)-\hat{L}_{i}\left(\mathbf{w}_{(k)}^{*}, I_{k}\right) \geq 0
\end{align}
\end{linenomath}
At the same time, we suppose: $\exists v_{k}$, 
\begin{linenomath}
\begin{align}
\label{exs}
    \hat{L}_{v_{k}}\left(\mathbf{w}_{(k)}, I_{k}\right)-\hat{L}_{v_{k}}\left(\mathbf{w}_{(k)}^{*}, I_{k}\right) \geq \frac{l_{k}}{2}\left\|\mathbf{w}_{(k)}-\mathbf{w}_{(k)}^{*}\right\|^{2}
\end{align}
\end{linenomath}
Now, let's handle $\|\mathbf{w}_{(k+1)}-\mathbf{w}^*_{(k+1)}\|^2$, from the definition of $\mathbf{w}^*_{(k+1)}$:
\begin{linenomath}
\begin{align}
\label{zong}
\nonumber\left\|\mathbf{w}_{(k+1)}-\mathbf{w}_{(k+1)}^{*}\right\|^{2} & \leq\left\|\mathbf{w}_{(k+1)}-\mathbf{w}_{(k)}^{*}\right\|^{2}=\left\|\mathbf{w}_{(k)}-\lambda_{k} \hat{g}_{k}+\eta_{k} \hat{d}_{k}-\mathbf{w}_{(k)}^{*}\right\|^{2} 
\\
&\nonumber=\left\|\mathbf{w}_{(k)}-\mathbf{w}_{(k)}^{*}\right\|^{2}-2\left\langle\mathbf{w}_{(k)}-\mathbf{w}_{(k)}^{*}, \lambda_{k} \hat{g}_{k}\right\rangle+
\\
&\quad\quad2\left\langle\mathbf{w}_{(k)}-\mathbf{w}_{(k)}^{*}, \eta_{k} \hat{d}_{k}\right\rangle+\left\|-\lambda_{k} \hat{g}_{k}+\eta_{k} \hat{d}_{k}\right\|^{2}
\end{align}
\end{linenomath}

Suppose we use the average parameter for the initial point of FedEcho, which means $\hat{g}_k$ is the weighted average parameter update of each client. Use $h_i$ to represent the weight coefficient, then for the second item, we have:
\begin{linenomath}
\begin{align}
\label{convex}
\left\langle\mathbf{w}_{(k)}^{*}-\mathbf{w}_{(k)}, \hat{g}_{k}\right\rangle \leq&  \sum_{i: I_{i}=1} h_{i}\left(L_{i}\left(\mathbf{w}_{(k)}^{*}\right)-L_{i}\left(\mathbf{w}_{(k)}\right)\right)-\frac{\sigma_{k}}{2}\left\|\mathbf{w}_{(k)}-\mathbf{w}_{(k)}^{*}\right\|^{2} \\
\nonumber=&-\sum_{i} h_{i}\left(\hat{L}_{i}\left(\mathbf{w}_{(k)}, I_{k}\right)-\widehat{L}_{i}\left(\mathbf{w}_{(k)}^{*}, I_{k}\right)\right)-\\&
\label{asss}
\frac{\sigma_{k}}{2}\left\|\mathbf{w}_{(k)}-\mathbf{w}_{(k)}^{*}\right\|^{2} \leq \frac{-h_{v_{k}} l_{k}-\sigma_{k}}{2}\left\|\mathbf{w}_{(k)}-\mathbf{w}_{(k)}^{*}\right\|^{2}
\end{align}
\end{linenomath}
where the Eq.\ref{convex} is from the $\sigma$ convex assumption, the Eq.\ref{asss} is from Eq.\ref{abs} and Eq.\ref{exs}. For the remaining items of Eq.\ref{zong}, we have: 
\begin{linenomath}
\begin{align}
\nonumber
\left\|-\lambda_{k} \hat{g}_{k}+\eta_{k} \hat{d}_{k}\right\|^{2} &=\left\|\lambda_{k} \hat{g}_{k}\right\|^{2}+\left\|\eta_{k} \hat{d}_{k}\right\|^{2}+2\left\langle-\lambda_{k} \hat{g}_{k}, \eta_{k} \hat{d}_{k}\right\rangle \\
\label{ml}
& \quad\quad\quad\leq \lambda_{k}^{2} M^{2}+\eta_{k}^{2} G^{2}+2\left\langle-\lambda_{k} \hat{g}_{k}, \eta_{k} \hat{d}_{k}\right\rangle \\
\label{item}
& \quad\quad\quad\leq \lambda_{k}^{2} M^{2}+\eta_{k}^{2} G^{2}+\lambda_{k}^{2} M^{2}+\eta_{k}^{2} G^{2} \\
\label{item2}
\left\langle \mathbf{w}_{(k)}-\mathbf{w}_{(k)}^{*}, \hat{d}_{k}\right\rangle & \leq \frac{1}{2}\left(\left\|\mathbf{w}_{(k)}-\mathbf{w}_{(k)}^{*}\right\|^{2}+\left\|\hat{d}_{k}\right\|^{2}\right)
\end{align}
\end{linenomath}
The Eq.\ref{ml} follows from the $M$-Lipschitz continuous assumption and the upper bound of $\hat{d}_k$. Take Eq.\ref{asss}, Eq.\ref{item} and Eq.\ref{item2} into Eq.\ref{zong}, we have:
\begin{linenomath}
\begin{align}
\label{abo}
    &\nonumber\left\|\mathbf{w}_{(k+1)}-\mathbf{w}_{(k+1)}^{*}\right\|^{2} \\&\leq\left(1-\left(h_{v_{k}} l_{k}+\sigma_{k}\right) \lambda_{k}+\eta_{k}\right)\left\|\mathbf{w}_{(k)}-\mathbf{w}_{(k)}^{*}\right\|^{2}+\eta_{k} G^{2}+2 \lambda_{k}^{2} M^{2}+2 \eta_{k}^{2} G^{2}
\end{align}
\end{linenomath}
suppose $h_{v_{k}} l_{k}+\sigma_{k} \geq c$ and  $\pi_{k}=\prod_{s=1}^{k}\left(1-c \lambda_{s}+\eta_{s}\right)$, from Eq.\ref{abo} we can get:
\begin{linenomath}
\begin{align}
\label{two}
\nonumber
    \mathbb{E}\left[\left\|\mathbf{w}_{(k+1)}-\mathbf{w}_{(k+1)}^{*}\right\|^{2}\right] \leq 
&\pi_{k}\left(1-c \lambda_{0}+\eta_{0}\right) \mathbb{E}\left[\left\|\mathbf{w}_{(0)}-\mathbf{w}_{(0)}^{*}\right\|^{2}\right]+
    \\&\sum_{s=0}^{k} \frac{\pi_{k}}{\pi_{s}}\left(\left(2 \eta_{s}^{2}+\eta_{s}\right) G^{2}+2 \lambda_{s}^{2} M^{2}\right)
\end{align}
\end{linenomath}
let $\lambda_{s}=\frac{3}{c(s+2)}$ and $\eta_{s}=\frac{1}{(s+2)(s+1)}$, then:
\begin{linenomath}
\begin{align}
\nonumber
&\pi_{k}=\prod_{s=1}^{k}\left(\frac{s^{2}}{(s+2)(s+1)}\right) 
\\&=\nonumber\frac{1 \times 1}{2 \times 3} \frac{2 \times 2}{3 \times 4} \frac{3 \times 3}{4 \times 5} \ldots \frac{(k-2)^{2}}{(k-1) k} \frac{(k-1)^{2}}{k(k+1)} \frac{k^{2}}{(k+1)(k+2)} \\
\label{kk}
&=\frac{2}{(k+1)^{2}(k+2)}
\end{align}
\end{linenomath}
we can see $\pi_k\to 0 $, the first item of Eq.\ref{two} approaches $0$ as $k$ increases. Take Eq.\ref{kk} into the second item of Eq.\ref{two}, we have:
\begin{linenomath}
\begin{align}
\nonumber\sum_{s=0}^{k} \frac{\pi_{k}}{\pi_{s}}\left(2 \eta_{s}^{2}+\eta_{s}\right)&=\frac{2}{(k+1)^{2}(k+2)}
\sum_{s=0}^{k}\left(\frac{1}{(s+2)}+\frac{s+1}{2}\right) \\
\nonumber
&\nonumber\leq \frac{2}{(k+1)^{2}(k+2)}\left((k+1)+\frac{(k+1)^{2}}{2}\right)\\\nonumber&=\frac{2}{(k+1)(k+2)}+\frac{1}{2(k+2)}\\
\nonumber
\sum_{s=0}^{k} \frac{\pi_{k}}{\pi_{s}}\left(2 \lambda_{s}^{2}\right)=&\frac{9}{c^{2}} \frac{2}{(k+1)^{2}(k+2)} \sum_{s=0}^{k} \frac{(s+1)^{2}}{(s+2)} \\\nonumber
&\leq \frac{9}{c^{2}} \frac{2}{(k+1)^{2}(k+2)} \frac{(k+1)^{2}(k+1)}{(k+2)} \leq \frac{18}{c^{2}(k+2)}
\end{align}

\end{linenomath}

Finally, we have $    \mathbb{E}\left[\left\|\mathbf{w}_{(k+1)}-\mathbf{w}_{(k+1)}^{*}\right\|^{2}\right]\to 0$.

\end{proof}

From Eq.\ref{the2}, one can see that, as the communication round $k$ increases, the right-hand side of the inequality approaches $0$, which means that the algorithm is convergent. For non-convex cases, such as image classification models, we will give empirical proofs of convergence in the experimental part.

\section{Discussions}
\label{diss}
In this paper, we propose a novel model aggregation method MA-Echo for one-shot FL. Compared to retraining a global model by Knowledge Distillation, MA-Echo does not need public data and has no training procedure. Compared with the traditional pure parameter aggregation methods such us OT~\cite{singh2020model}, MA-Echo makes the first attempt to utilize auxiliary the null-space projection matrices to aggregate model parameters, which clearly improves the performance. 
In the following we discuss three points that may receive attention:
\begin{itemize}
\item \textbf{The situations under which MA-Echo may not work.} When using the projection matrix $P$ to to build the objective Eq.\ref{mom2}, the input feature $\mathbf{x}$ and parameter $\mathbf{w}$ have the same dimensionality $d$ by default. In this case, if the rank of the feature subspace is close to $d$, then the rank of the null space of feature is close to $0$, so it is difficult to find an orthogonal direction to make $\mathbf{x}^\top(\mathbf{w}-\mathbf{w}_i)=0$. A potential solution is to flexibly increase the dimensionality of $\mathbf{w}$ or perform dimensionality reduction on the data matrix $X$, so that there can be more degrees of freedom to find the orthogonal direction (not necessarily strictly orthogonal, but can minimize the loss in Eq.\ref{mom2} ).

\item {\textbf{Overhead in computation and communication.} We discuss the calculation cost (including the training cost) and transmission cost respectively: 1) For the training cost, the projection matrix calculation only requires an additional epoch model forward propagation. The calculation cost is less than the cost of one epoch of training. Compared with the client-side overall training, we believe the cost is acceptable. 2) For the calculation cost in aggregation, we will show the elapsed time of all methods in the experiment. Compared with distillation based aggregation algorithm DENSE, our method only consumes a small amount of calculation cost. 3) For the transmission cost.  For the $i$-th layer of a neural network, the parameter is $W^l \in \mathbb{R}^{C_{in} \times C_{out}}$, where $C_{in}$ and $C_{out}$ is the dimension of the input and output feature. The projection matrix corresponding to $W^l$ is $P^l \in \mathbb{R}^{C_{in} \times C_{in}}$. The communication cost of the projection matrix is $\frac{ C_{in}}{ C_{out}}$ times the parameters.  We will also conduct additional experiments to show that we can easily reduce the size of the projection matrix by SVD decomposition without affecting the performance. }
\item \textbf{Privacy.} The projection matrix can be regarded as a special network layer whose input is not data, but model parameters. Therefore, compared to uploading model parameters directly, the projection matrix does not bring more privacy leakage.
\end{itemize}

\begin{figure}[t]
\centering
    \subfloat[$\beta=0.01$]{
    \includegraphics[width=0.25\linewidth]{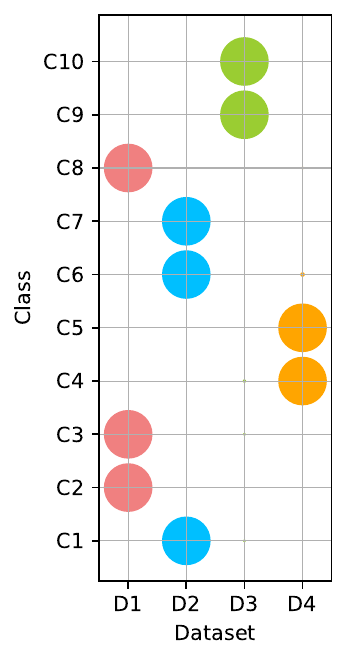}
    }
    \subfloat[$\beta=0.95$]{
    \includegraphics[width=0.25\linewidth]{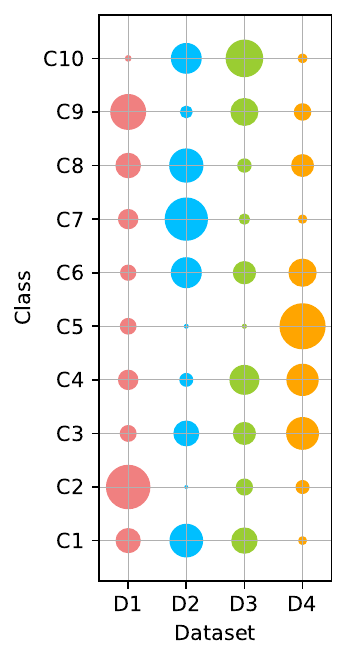}
    }
    \subfloat[$\beta=20$]{
    \includegraphics[width=0.25\linewidth]{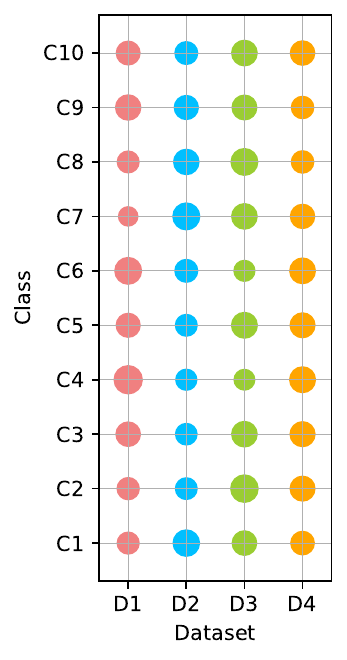}
    }

\caption{Visualization of data partition (Better viewed in color). (a) $\beta=0.01$, the labels in the trainset of different local models hardly overlap. (c) Different local models have similar data distribution.}
\label{partition}
\end{figure}
\begin{figure*}[ht]

\centering
\subfloat[Diff-init 2-MLPs]{
\includegraphics[width=0.235\linewidth]{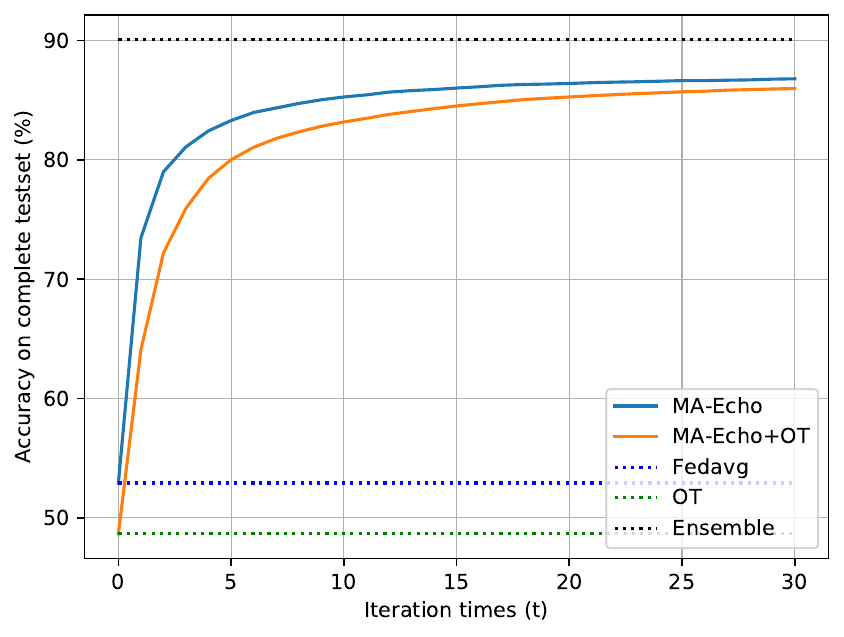}
}
\subfloat[Diff-init 5-MLPs]{
\includegraphics[width=0.235\linewidth]{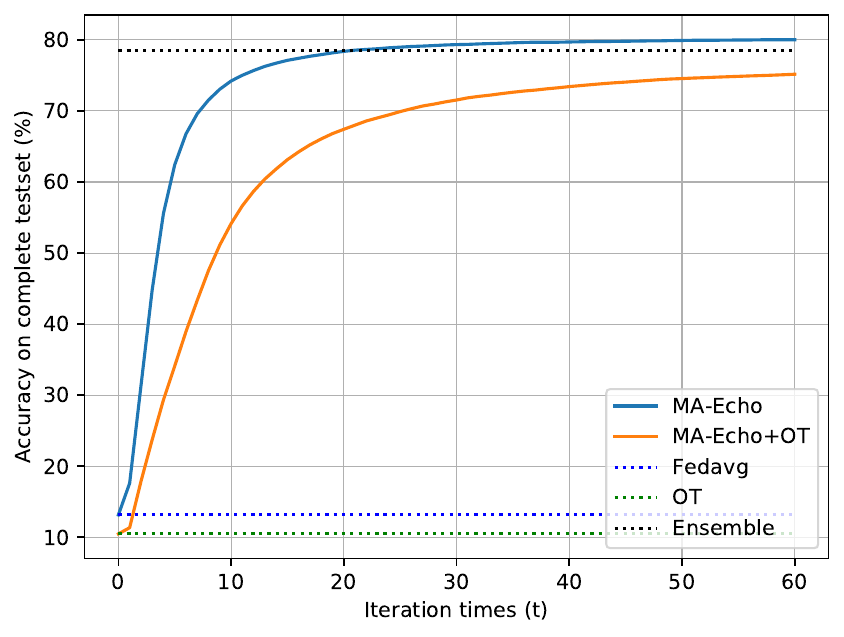}
}
\subfloat[Diff-init 2-CNNs]{
\includegraphics[width=0.235\linewidth]{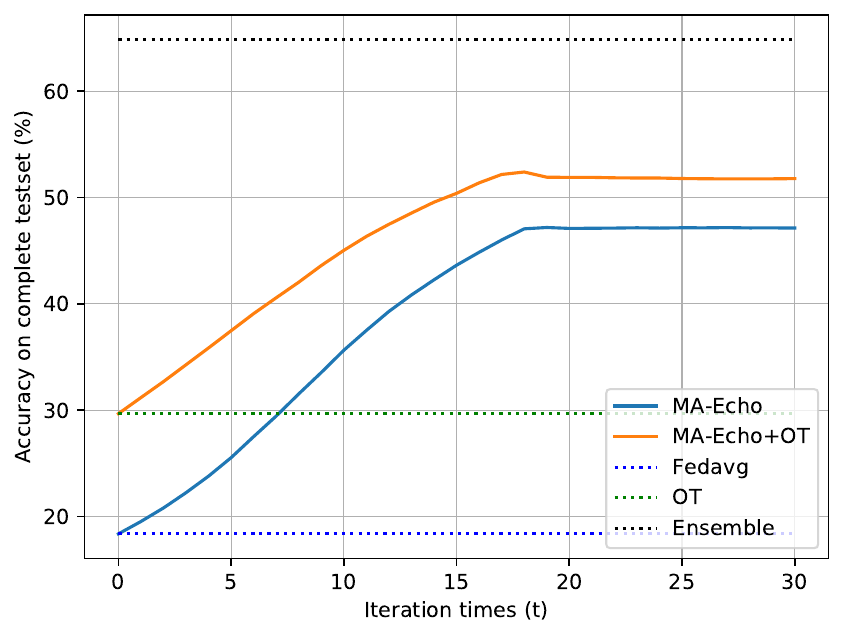}
}
\subfloat[Diff-init 5-CNNs]{
\includegraphics[width=0.235\linewidth]{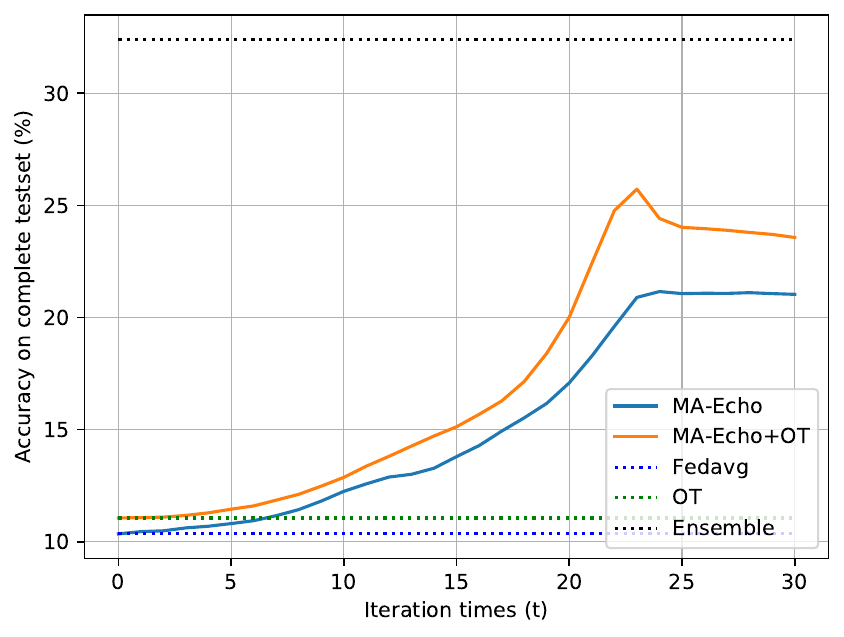}
}
\quad
\subfloat[Same-init 2-MLPs]{
\includegraphics[width=0.235\linewidth]{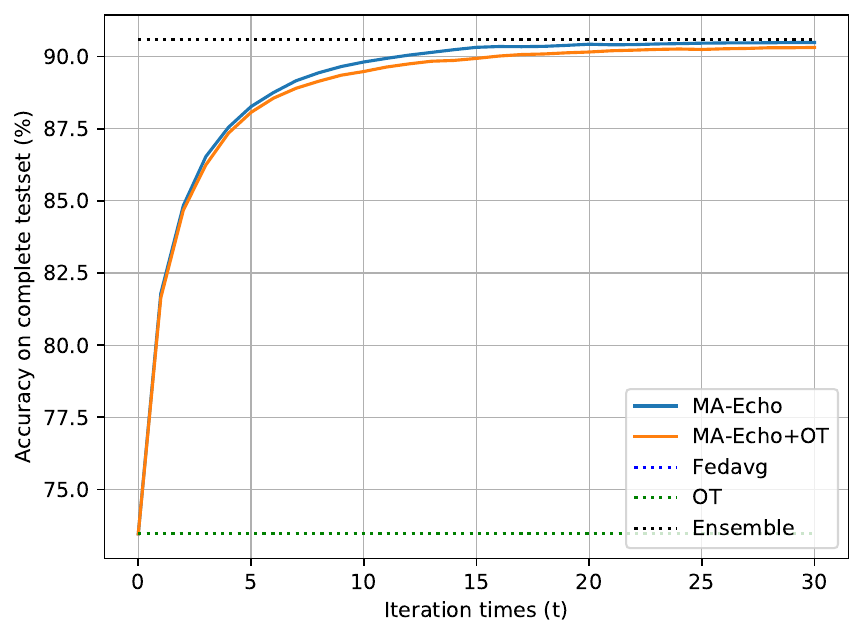}
}
\subfloat[Same-init 5-MLPs]{
\includegraphics[width=0.235\linewidth]{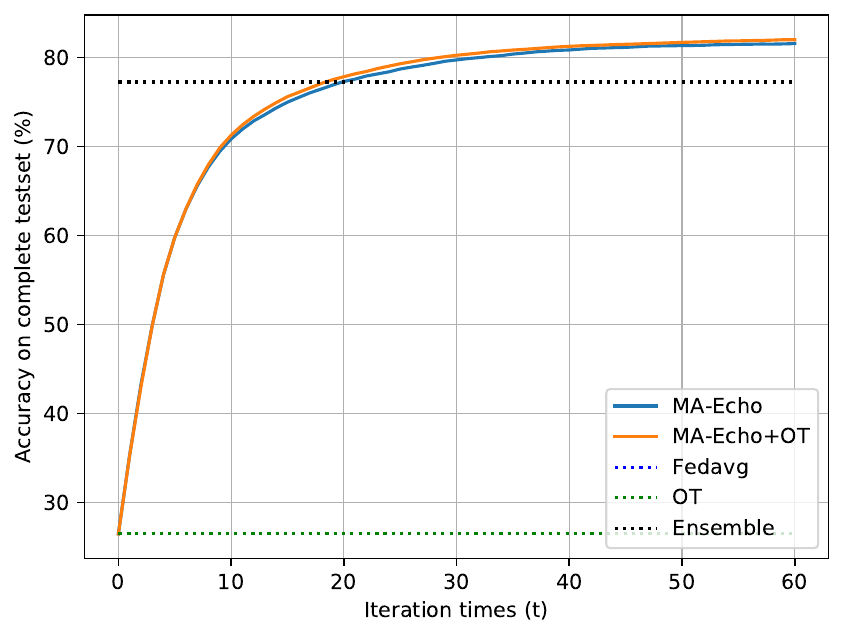}
}
\subfloat[Same-init 2-CNNs]{
\includegraphics[width=0.235\linewidth]{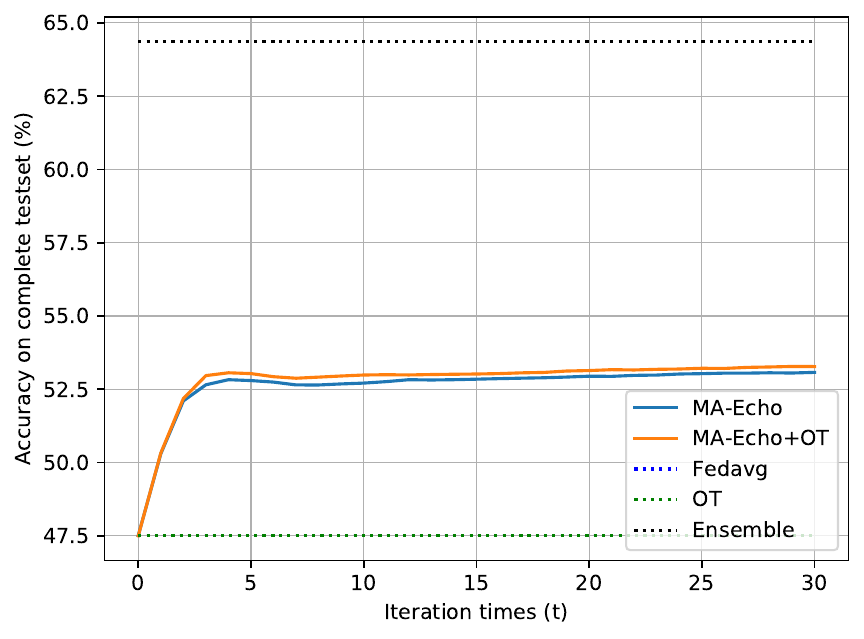}
}
\subfloat[Same-init 5-CNNs]{
\includegraphics[width=0.235\linewidth]{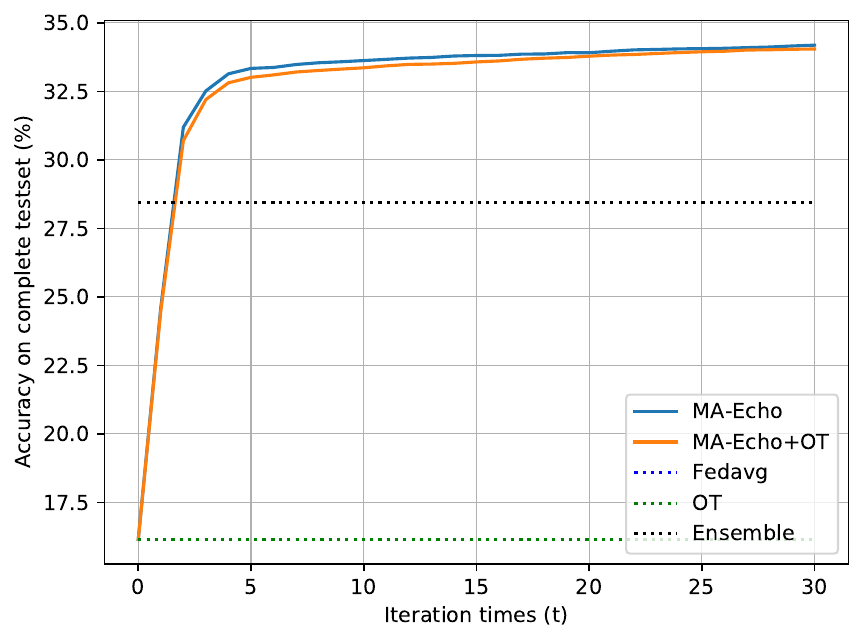}
}

\label{agr_results}
\caption{
The performance of model aggregation in one-shot FL (Better viewed in color). The results of FedAvg, OT, and Ensemble are fixed values because they have no iterative procedure. `Diff-init 2-MLPs' means that the two local models have different initialization before training. MA-Echo outperforms its competitors and is even better than Ensemble in some situations. For (c) and (d), we use $\text{Norm}(
W)= \text{torch.norm}(W,\text{dim}=1)$ in Algotithm~\ref{algo1}.
}
\label{agr_exp}
\end{figure*}

\begin{table*}[t] 
\center
\caption{{Multi-model one-shot aggregation.}}
\label{table_mt}
\resizebox{0.9\textwidth}{!}{%
\begin{tabular}{ccccccc}
\Xhline{1.2pt}
\multicolumn{1}{c|}{}                & Local acc & Average & OT         & DENSE      & Ours  & Ensemble \\ \hline
\multicolumn{7}{c}{5 clients}                                                                           \\ \hline
\multicolumn{1}{c|}{$\beta $= 0.01}    & 24.65     & 20.97   & 20.97      & 32.77      & \textbf{80.31} & 50.50    \\
\multicolumn{1}{c|}{$\beta $ = 0.1}     & 39.27     & 34.41   & 34.41      & 46.29      & \textbf{74.26} & 56.50    \\
\multicolumn{1}{c|}{$\beta $ = 0.5}     & 68.66     & 64.94   & 64.94      & 56.67      & \textbf{78.34} & 75.65    \\\hline
\multicolumn{1}{c|}{elapsed time (s)} & \textbackslash       & 0.003   & 0.117      & 565.240    & 1.045 & \textbackslash      \\ \hline
\multicolumn{7}{c}{10 clients}                                                                          \\ \hline
\multicolumn{1}{c|}{$\beta $= 0.01}    & 11.86     & 17.61   & 17.63      & 26.01      & \textbf{79.30} & 37.71    \\
\multicolumn{1}{c|}{$\beta $ = 0.1}     & 35.13     & 35.41   & 35.41      & 33.51      & \textbf{74.07} & 52.12    \\
\multicolumn{1}{c|}{$\beta $= 0.5}     & 61.37     & 59.75   & 59.75      & 64.64      & \textbf{81.80} & 76.61    \\\hline
\multicolumn{1}{c|}{elapsed time  (s)}    &\textbackslash        & 0.005   & 0.233      & 626.930    & 1.883 & \textbackslash      \\ \hline
\multicolumn{7}{c}{20 clients}                                                                          \\ \hline
\multicolumn{1}{c|}{$\beta $ = 0.01}    & 11.64     & 18.78   & 18.77      & 35.50      & \textbf{78.70} & 50.03    \\
\multicolumn{1}{c|}{$\beta $ = 0.1}     & 21.80     & 37.00   & 36.99      & 37.35      & \textbf{80.40} & 60.02    \\
\multicolumn{1}{c|}{$\beta $ = 0.5}     & 52.22     & 66.76   & 66.76      & 57.77      & \textbf{83.58} & 75.73    \\\hline
\multicolumn{1}{c|}{elapsed time  (s)}    & \textbackslash        & 0.008   & 0.557 & 760.556 & 3.299 &\textbackslash      \\ \hline
\multicolumn{7}{c}{50 clients}                                                                          \\ \hline
\multicolumn{1}{c|}{$\beta $ = 0.1}     & 18.77     & 28.835  & 28.83      & 47.345     & \textbf{75.3}  & 50.22    \\
\multicolumn{1}{c|}{$\beta $ = 0.5}     & 37.14     & 56.02   & 56.02      & 42.7       & \textbf{79.37} & 68.535   \\
\multicolumn{1}{c|}{$\beta $ = 10}      & 64.08     & 71.805  & 71.8       & 52.39      & \textbf{75.27} & 72.905   \\\hline
\multicolumn{1}{c|}{elapsed time  (s)}    & \textbackslash        & 0.010    & 1.473      & 1270.330    & 9.603  & \textbackslash     \\ \Xhline{1.2pt}
\end{tabular}
}
\end{table*}

\begin{table}[t]
\center
\caption{{Aggregation performance on CIFAR100.}}
\label{table_c100}
\resizebox{0.8\textwidth}{!}{%
\begin{tabular}{cccccc}
\Xhline{1.2pt}
\multicolumn{1}{c|}{}             & Local acc & Average & OT    & Ours  & Ensemble \\ \hline
\multicolumn{6}{c}{5 clients}                                                      \\ \hline
\multicolumn{1}{c|}{$\beta$ = 0.01} & 16.91     & 18.20  & 18.20 & 25.26 & 22.95    \\
\multicolumn{1}{c|}{$\beta$ = 0.1}  & 24.27     & 20.68   & 24.86 & 29.10 & 28.15    \\
\multicolumn{1}{c|}{$\beta$ = 0.5}  & 36.29     & 41.81   & 41.81 & 45.94 & 46.98    \\ \hline
\multicolumn{6}{c}{10 clients}                                                     \\ \hline
\multicolumn{1}{c|}{$\beta$ = 0.01} & 10.10     & 9.18    & 9.18  & 16.81 & 16.11    \\
\multicolumn{1}{c|}{$\beta$ = 0.1}  & 19.12     & 21.16   & 21.17 & 27.13 & 27.36    \\
\multicolumn{1}{c|}{$\beta$ = 0.5}  & 32.42     & 39.05   & 39.05 & 43.26 & 45.00    \\ \hline
\multicolumn{6}{c}{20 clients}                                                     \\ \hline
\multicolumn{1}{c|}{$\beta$ = 0.01} & 6.44      & 4.59    & 4.59  & 7.88  & 8.20     \\
\multicolumn{1}{c|}{$\beta$ = 0.1}  & 14.56     & 16.88   & 16.88 & 22.04 & 22.74    \\
\multicolumn{1}{c|}{$\beta$ = 0.5}  & 26.09     & 34.14   & 34.14 & 39.40 & 41.74    \\ \Xhline{1.2pt}
\end{tabular}
}
\end{table}

\begin{figure*}[ht]
\centering
\includegraphics[width=0.96\linewidth]{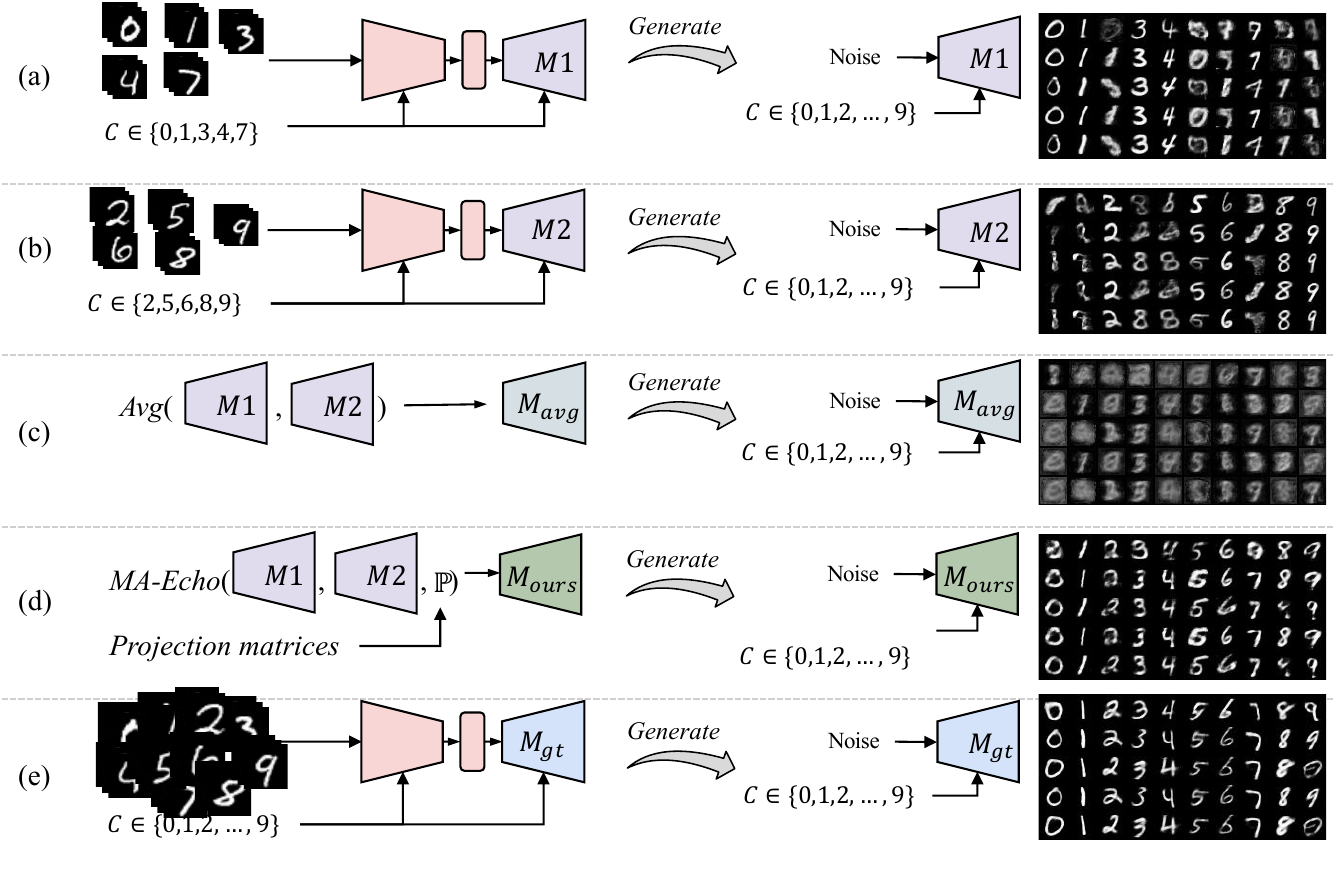}


\caption{
Images generated by five different decoders: (a) Model1: the decoder trained by $\{0,1,3,4,7\}$ digits; (b) Model2: the decoder trained by $\{2,5,6,8,9\}$ digits; (c) Average: average the two trained decoders; (d) Ours: the decoder aggregated by MA-Echo (e) GT decoder: a model trained by the whole MNIST dataset. It can be seen that through our aggregation method, the aggregation model can obtain the knowledge of Model1 and Model2 at the same time. Better viewed in color.
}
\label{cvae}
\end{figure*}

\begin{table}[t]  
\center
\caption{{Aggregation under the data heterogeneity caused by domain feature shift.}}
\label{table_ds}
\resizebox{0.7\textwidth}{!}{%
\begin{tabular}{cccccc}
\Xhline{1.2pt}
FEMNIST    & Local acc & Average & OT    & Ours  & ensemble \\ \hline
10 clients & 92.28     & 92.76   & 92.77 & 92.81 & 92.88    \\
50 clients & 80.05     & 81.8    & 81.79 & 82.26 & 82.22    \\ \hline
DomainNet  & Local acc & Average & OT    & Ours  & ensemble \\ \hline
6 clients  & 20.42     & 0.35    & 0.35  & 30.5  & 35.02    \\
12 clients & 18.49     & 0.17    & 0.17  & 30.06 & 35.01    \\ \Xhline{1.2pt}
\end{tabular}}
\end{table}

\begin{figure*}[ht]
\centering
\subfloat[Original ]{
\label{tsne_ori}
\includegraphics[width=0.235\linewidth]{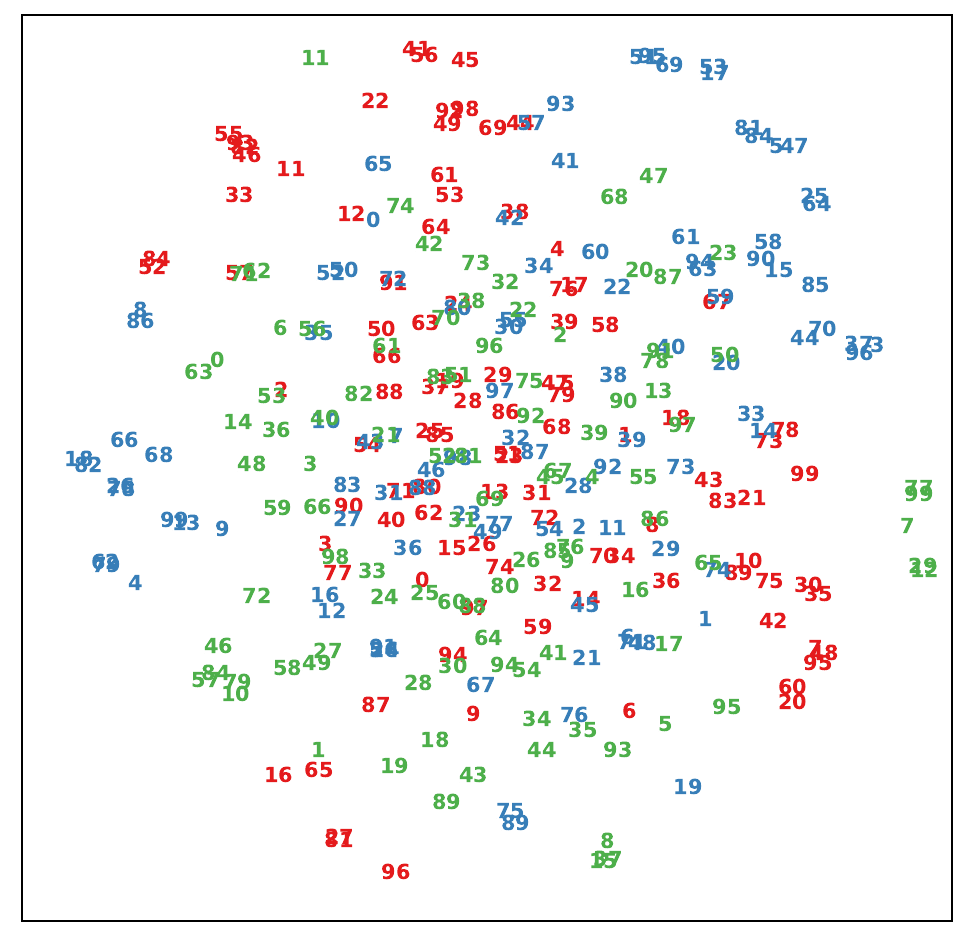}
}
\subfloat[OT ]{
\label{tsne_ot}
\includegraphics[width=0.235\linewidth]{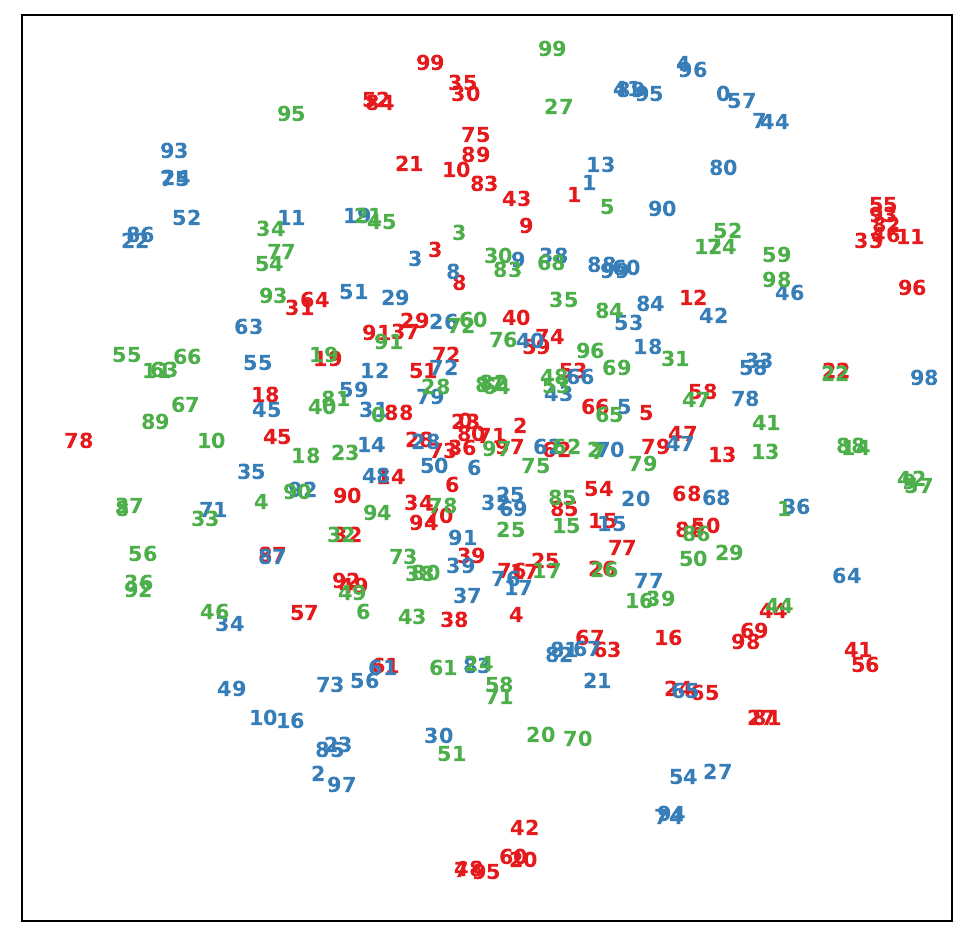}
}
\subfloat[Ours $1$th iter ]{
\label{tsne_ours1}
\includegraphics[width=0.235\linewidth]{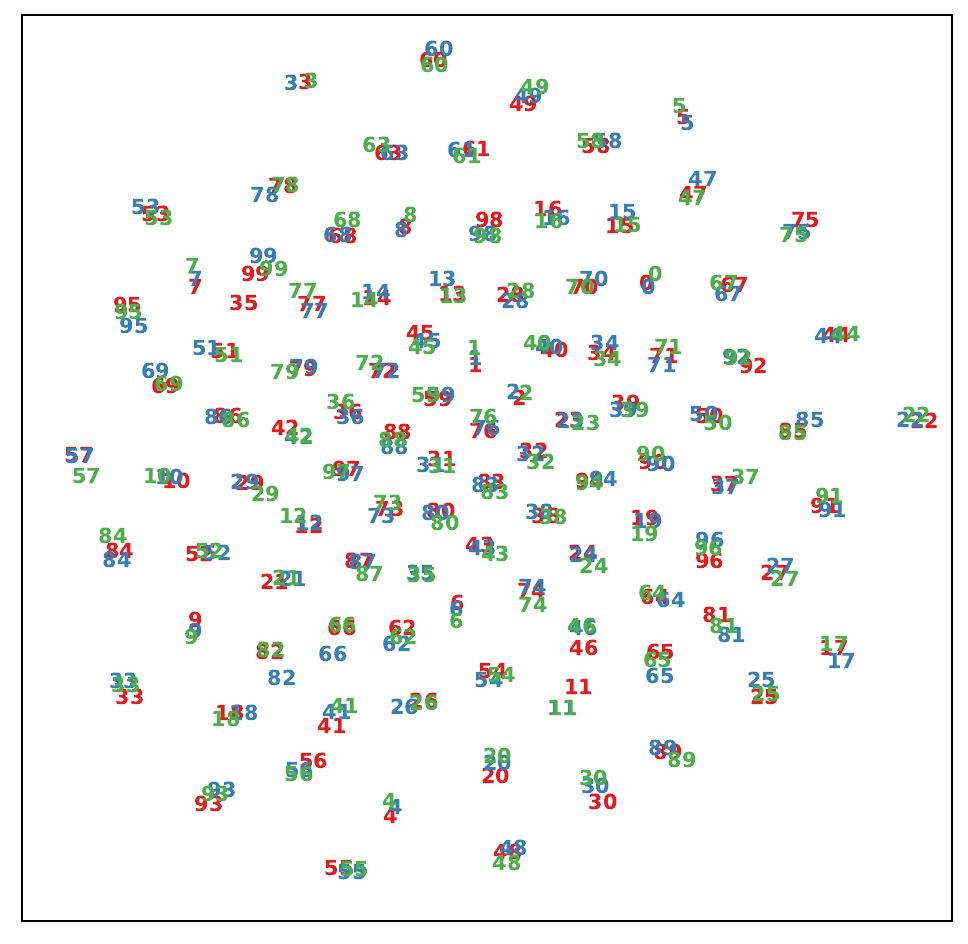}
}
\subfloat[Ours $30$th iter]{
\label{tsne_ours30}
\includegraphics[width=0.235\linewidth]{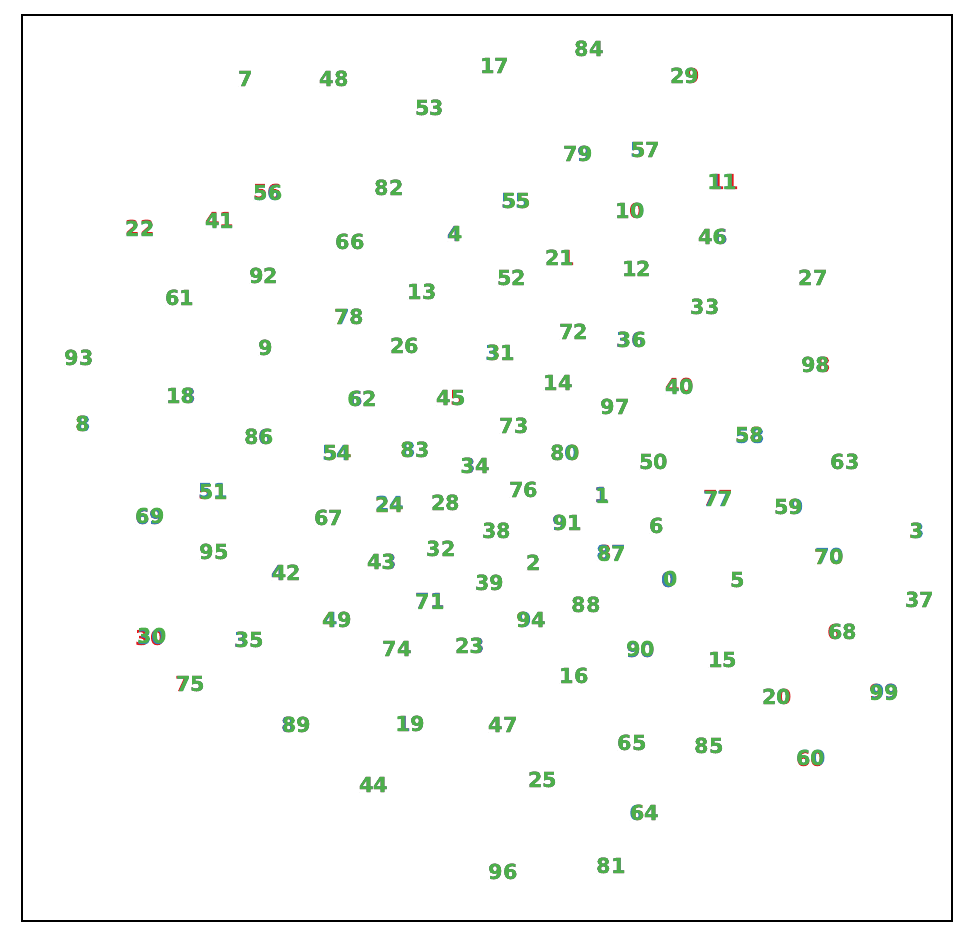}
}
\caption{
The visualization of parameter vectors in third layer in three-MLP aggregation, the numbers in parentheses are the average accuracy of the three local models in their local training data. Each number represents a row vector of parameter matrix $W_i$ of the third layer (the shape of $W_i$ is $100\times 200$), and each color corresponds to a local model. (a) Since the original models have different initial parameters, the three layers do not match well after the training is completed. (b) After rearranged by OT, some vectors can be matched. (c)$\sim$(d) After one iteration, MA-Echo can match most of the three models' neurons. After $30$ iterations, MA-Echo achieves a perfect match and does not weaken the accuracy too much. Better viewed in color.
}
\label{t_sne}
\end{figure*}





\begin{figure}[t]
\centering
    \subfloat[]{
    \label{sur}
    \includegraphics[width=0.46\linewidth]{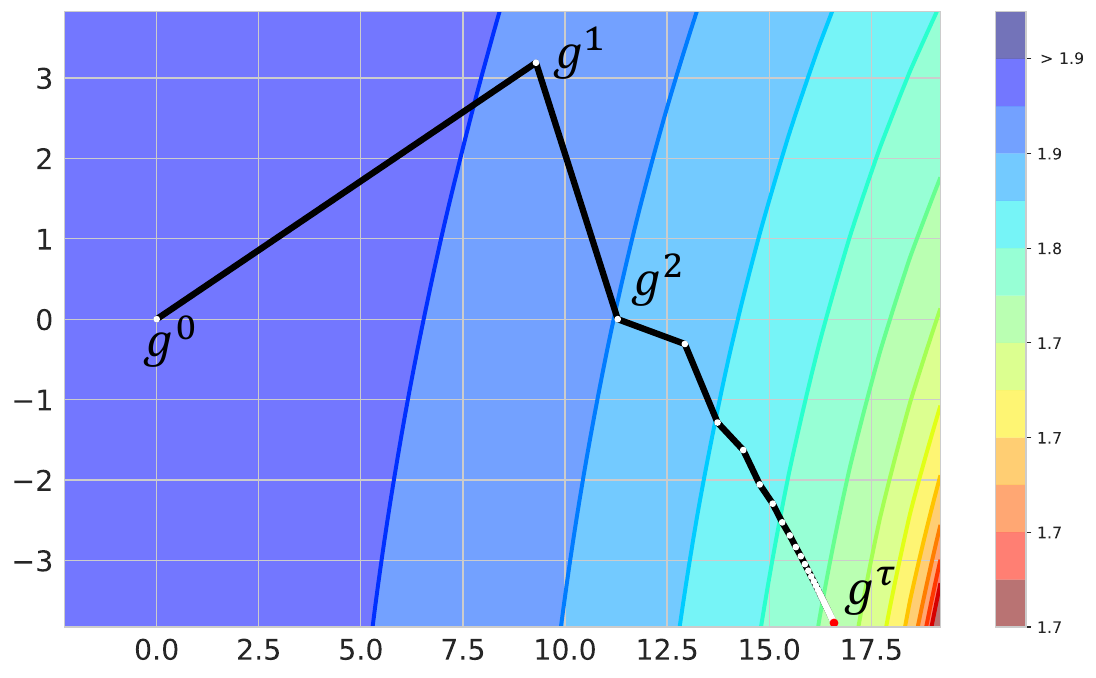}
    }
    \subfloat[]{
    \label{diff}
    \includegraphics[width=0.38\linewidth]{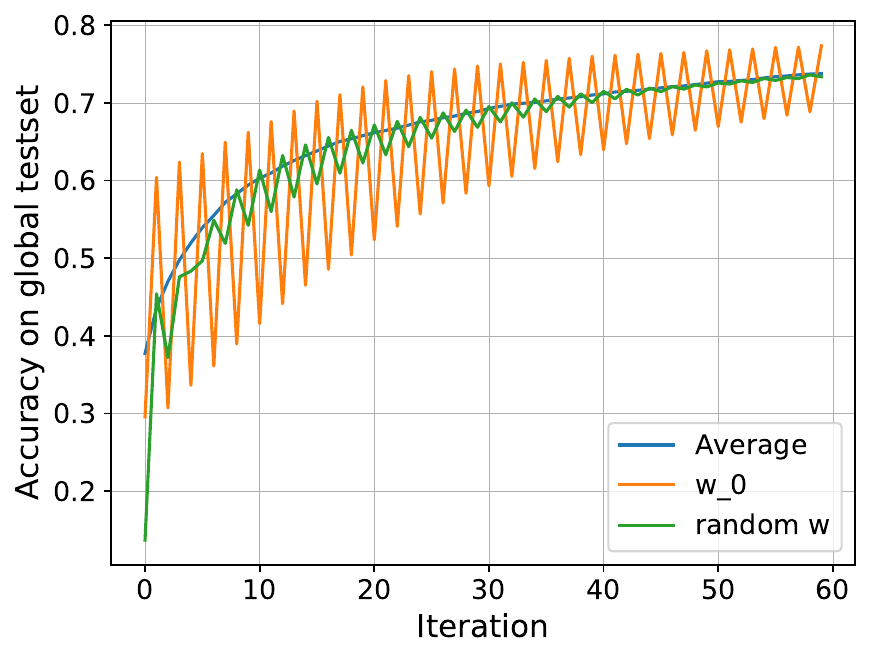}
    }

\caption{(a) The visualization of the iterations. $g^t$ is the projection of $\text{Flatten}(\{W^{1(t)},\cdots,W^{L(t)}\})$ on this 2D plane. It shows that MA-Echo keeps exploring the lower loss area in the parameter space. (b) Different initialization for $W^{l(0)}$ in Algorithm~\ref{algo1}.  $\textbf{Average}$: use $\bar{W}^{l(0)}=1/N\sum_i W_i^l$ for initialization. $\textbf{W}_0$: use one local model $W_0$ as initialization. $\textbf{random W}$: use random initialization. Better viewed in color.}
\end{figure}

\begin{table*}[tb]
\caption{Performance under varying degrees of non-identicalness of training data in two-MLP aggregation on MNIST and two-CNN aggregation on CIFAR-10. The smaller the $\beta$, the greater the degree of non-identicalness. When $\beta$ is close to 0, the support categories of each local model have no overlaps. MA-Echo works best among four compared model aggregation methods (first four methods) and can greatly improve the performance of OT. }
\resizebox{\textwidth}{!}{%

\begin{tabular}{c|c|c|c|c|c|c|c|c|c|c|c|c}
\Xhline{1.2pt}
\multicolumn{1}{c|}{{ }}  &\multicolumn{6}{c|}{{ MLP diff-init}} &\multicolumn{6}{c}{{ MLP same-init}}\\ \cline{1-7} \cline{8-13}
  $\beta$  & Fedavg  & PFNM   & OT     & MA-Echo &   Ensemble      & MA-Echo+OT        & Fedavg  & PFNM & OT  & MA-Echo  & Ensemble & MA-Echo+OT\\ \cline{1-7} \cline{8-13} 
 0.01 &    38.05  &  26.10   &  41.51 &   \textbf{84.49}  &  \textbf{\textit{89.80}}  & \textbf{\textit{86.78}}  &  66.80 &  35.21   &   66.80  & \textbf{89.50} & \textbf{\textit{90.95}} &  \textbf{\textit{89.31}}     \\
 0.50 & 62.15 &  54.15   &   76.40 &  \textbf{88.51}  & \textbf{\textit{93.78}} &   \textbf{\textit{89.34}}      & 75.20 &  \textbf{88.97}  & 75.20 &   88.71 &    \textbf{\textit{94.24}}    &  \textbf{\textit{88.92}}  \\
 1.50 & 70.23 & 54.30  &   86.99  &   \textbf{89.58}  &  \textbf{\textit{96.55}}& \textbf{\textit{93.43}}    &  84.60  &  85.49 & 84.60  &   \textbf{92.13}  &  \textbf{\textit{96.67}}  &   \textbf{\textit{91.87}}  \\

 20.0 &69.41 &  65.28  &   \textbf{92.71} &   92.31 &  \textbf{\textit{96.94}} & \textbf{\textit{95.49}}    & 96.78 & 95.93  & 96.78 &   \textbf{96.80}   &  \textbf{\textit{96.87}}  &    \textbf{\textit{96.76}}  \\
\hline
\hline
\multicolumn{1}{c|}{{ }}  &\multicolumn{6}{c|}{{ CNN diff-init}} &\multicolumn{6}{c}{{ CNN same-init}}\\ \cline{1-7} \cline{8-13}
   $\beta$ & Fedavg  & PFNM   & OT     & MA-Echo &   Ensemble      & MA-Echo+OT        & Fedavg  & PFNM & OT  & MA-Echo  & Ensemble & MA-Echo+OT\\ \cline{1-7} \cline{8-13} 
 0.01 &15.41  &    11.04   &   28.42 &  \textbf{ 41.53}  &  \textbf{\textit{60.59}}  &  \textbf{\textit{47.26}}     &  50.97 &   13.52  & 50.98  & \textbf{55.85}  &  \textbf{\textit{62.56}}   &  \textbf{\textit{56.18}}   \\
 0.50 & 19.61 &   14.78  &   45.06 &   \textbf{50.00}    &  \textbf{\textit{65.72}}  &\textbf{\textit{57.07}}     &   56.83 &   22.35   & 56.83 &   \textbf{60.24} &   \textbf{\textit{64.96}}    & \textbf{\textit{60.09}}  \\
 1.50 & 20.20 &   14.12    &   38.09  &   \textbf{47.37}  & \textbf{\textit{66.01}} & \textbf{\textit{53.31}}   &   62.86  &  29.24   & 62.86  &   \textbf{63.16}  &  \textbf{\textit{68.01}} & \textbf{\textit{63.09}}   \\
  20.0 &  29.16 &  12.21   &  54.16 &   \textbf{56.06} & \textbf{\textit{72.95}}  & \textbf{\textit{63.36}}   &  71.43 &   36.71   & \textbf{71.43} &   71.41    &   \textbf{\textit{73.04}}  &  \textbf{\textit{71.30}}   \\ 
\Xhline{1.2pt}
\end{tabular}%
}
\label{table_noniid}
\end{table*}
 
\begin{table}[ht]
\centering
\caption{The influence of the number of local training SGD steps.}
\resizebox{0.56\textwidth}{!}{%
\begin{tabular}{l|l|l |l |l |l |l |l |l |l |l }
\Xhline{1.2pt}
                                                                                        init & \#Steps & M1   & M2   & M3   & M4   & M5   & Avg  & OT   & Ours          & Ens.          \\ \cline{1-11} 
                                                                                         & 20     & 12.5 & 14.0 & 13.3 & 17.6 & 15.2 & 10.8 & 10.5 & \textbf{39.2} & \textit{12.3} \\
                                                                                         & 50     & 14.3 & 19.6 & 18.1 & 19.2 & 19.5 & 12.8 & 12.1 & \textbf{53.7} & \textit{33.4} \\ 
                                                                                         & 100    & 16.2 & 21.2 & 19.8 & 22.2 & 21.7 & 12.3 & 11.5 & \textbf{63.3} & \textit{49.3} \\
\multirow{-4}{*}{\begin{tabular}[c]{@{}l@{}}Diff\\ \end{tabular}} & 500    & 17.9 & 23.0 & 23.8 & 27.5 & 22.1 & 12.0 & 12.9 & \textbf{70.0} & \textit{63.0} \\ \hline
                                                                                         & 20     & 11.3 & 11.0 & 15.3 & 14.4 & 16.1 & 11.4 & 11.4 & \textbf{17.9} & \textit{10.2} \\ 
                                                                                         & 50     & 14.4 & 20.3 & 18.9 & 19.4 & 19.3 & 14.6 & 14.6 & \textbf{59.4} & \textit{34.8} \\
                                                                                         & 100    & 16.2 & 21.1 & 20.1 & 22.3 & 21.8 & 21.7 & 21.7 & \textbf{70.1} & \textit{49.9} \\ 
\multirow{-4}{*}{Same}                                                              & 500    & 17.9 & 23.2 & 23.9 & 27.5 & 22.1 & 24.1 & 24.1 & \textbf{76.5} & \textit{64.0} \\ 
\Xhline{1.2pt}
\end{tabular}%
}
\label{table_local}
\end{table}

\begin{figure}[t]
\centering
    \subfloat[]{
    \includegraphics[width=0.38\linewidth]{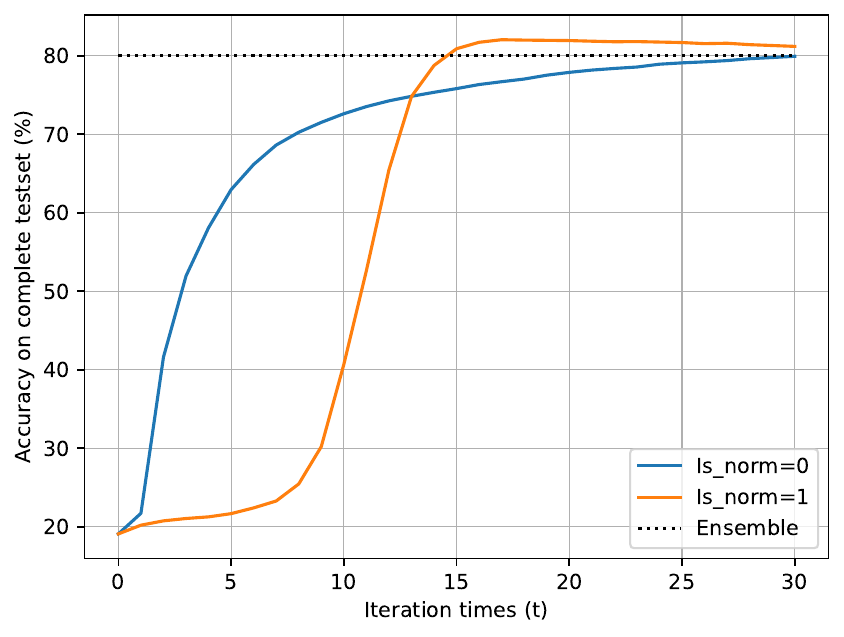}
    }
    \subfloat[]{
    \includegraphics[width=0.38\linewidth]{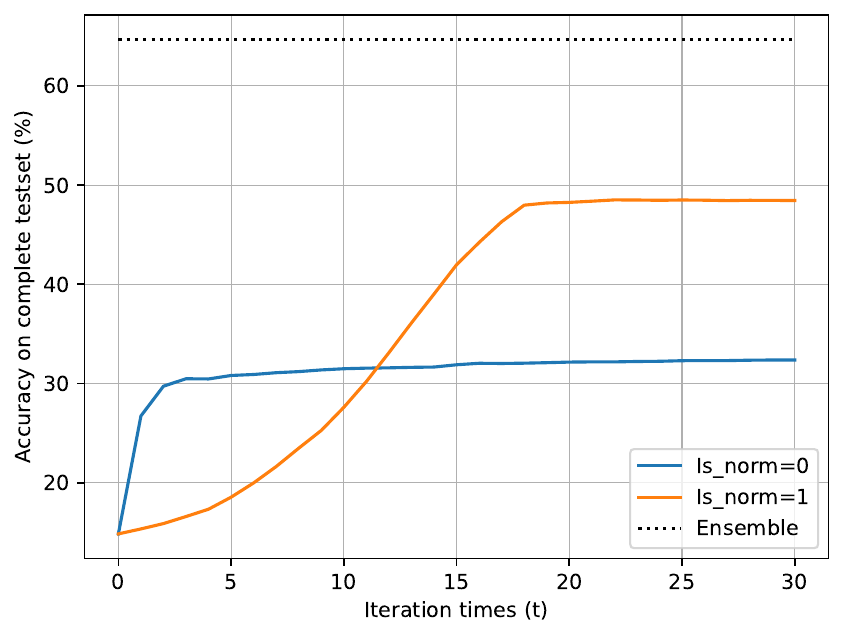}
    }

\caption{The influence of normalization, ``Is-norm'' means using $\text{Norm}(\cdot)$. (a): For five MLPs aggregation. Convergence is faster when $\text{Is-Norm}=1$, because we can use a larger step size $\lambda$. (b): Two CNNs aggregation, $\text{Is-Norm}=1$ can bring a significant improvement. Better viewed in color.}
\label{norm}
\end{figure}

\begin{figure*}[tb]
\centering
\subfloat[Global Model accuracy]{
\includegraphics[width=0.30
\linewidth]{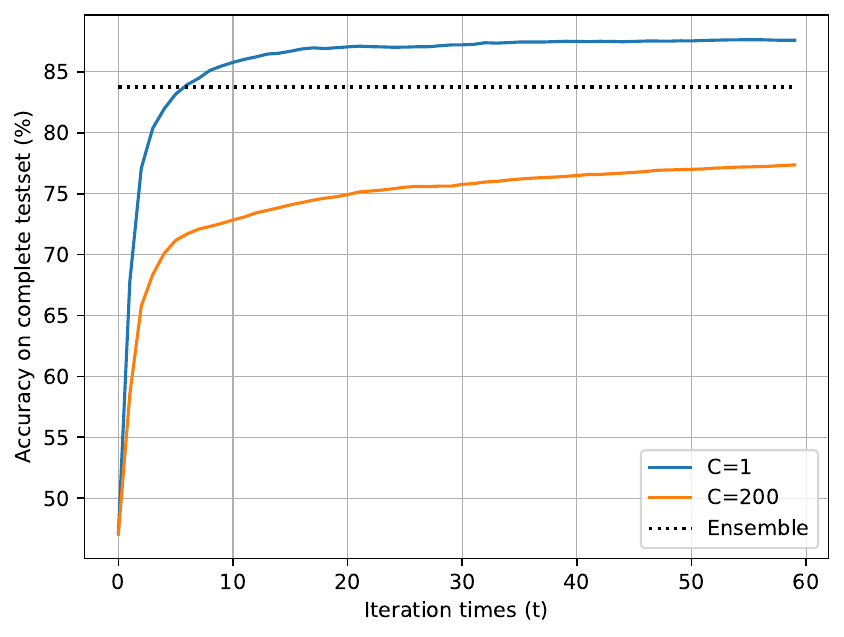}
}%
\subfloat[Accuracy decay in $\mu=1$]{
\includegraphics[width=0.31\linewidth]{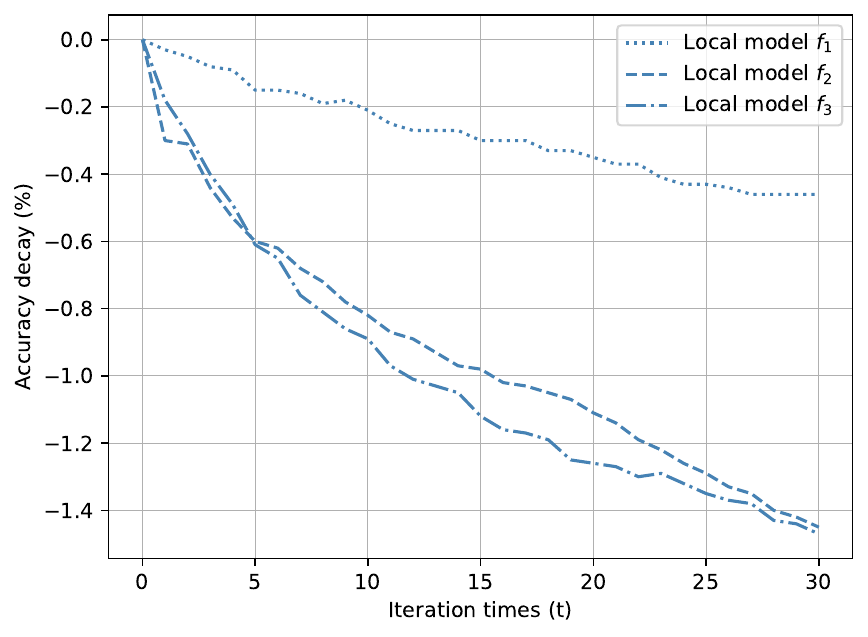}
}%
\subfloat[Accuracy decay in $\mu=200$]{
\includegraphics[width=0.32\linewidth]{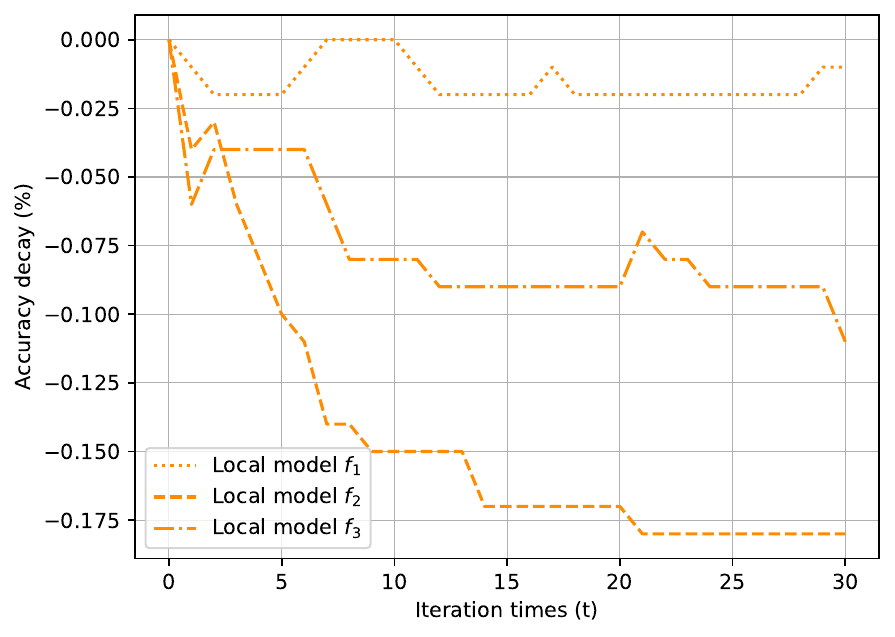}
}%
\centering
\caption{
The impact of penalty coefficient $\mu$ in Eq.\ref{optv}. (a)  Under the setting of $\mu=1$, the aggregation result for 3-MLPs is better than $\mu=200$. It shows the great effect of relaxation. (b) and (c)  Local models' accuracy decreases during iterative optimization. A smaller $\mu$ can bring appropriate relaxation to the projection constraint to improve the accuracy of the aggregated global model, at a cost of slight performance loss on local models. Better viewed in color.}
\label{penltyC}
\end{figure*}

\begin{table}[t]  
\center
\caption{{The change of communication size and aggregation performance after SVD compression.  \#params represents the total parameter amount of all four projection matrices.}}
\label{table_svd}
\resizebox{0.7\textwidth}{!}{%
\begin{tabular}{cccc|c|c}
\Xhline{1.2pt}
\multicolumn{4}{c|}{Number of principal components} & \multirow{2}{*}{\#params (M)} & \multirow{2}{*}{Acc} \\ \cline{1-4}
layer1      & layer2      & layer3     & layer4     &                               &                      \\ \hline
784         & 400         & 200        & 100        & 0.824656                      & 81.22                \\
200         & 100         & 50         & 30         & 0.2098                        & 80.64                \\
20          & 20          & 10         & 10         & 0.02668                       & 80.55                \\
5           & 5           & 5          & 5          & 0.00742                       & 80.48                \\
2           & 2           & 2          & 2          & 0.002968                      & 79.65                \\
1           & 1           & 1          & 1          & 0.001484                      & 76.86                \\ \Xhline{1.2pt}
\end{tabular}}
\end{table}

\begin{figure*}[th]
 
\centering
\subfloat[MNIST,$\frac{m}{N}$=$\frac{10}{100}$,$\#Class$=$1$]{
\includegraphics[width=0.38
\linewidth]{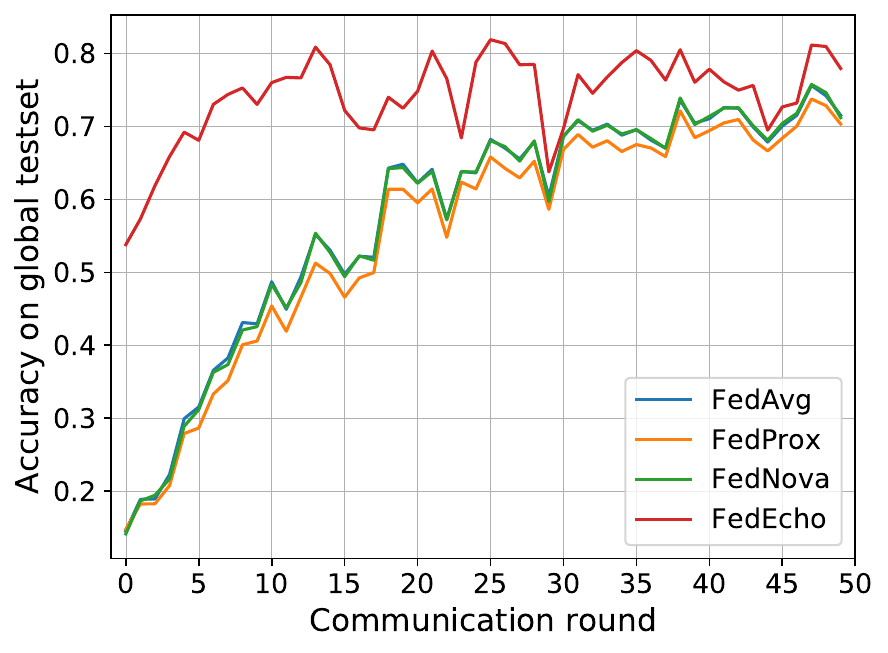}
}%
\subfloat[FEMNIST,$\frac{m}{N}$=$\frac{20}{1000}$,$\#Class$=$1$]{
\includegraphics[width=0.38\linewidth]{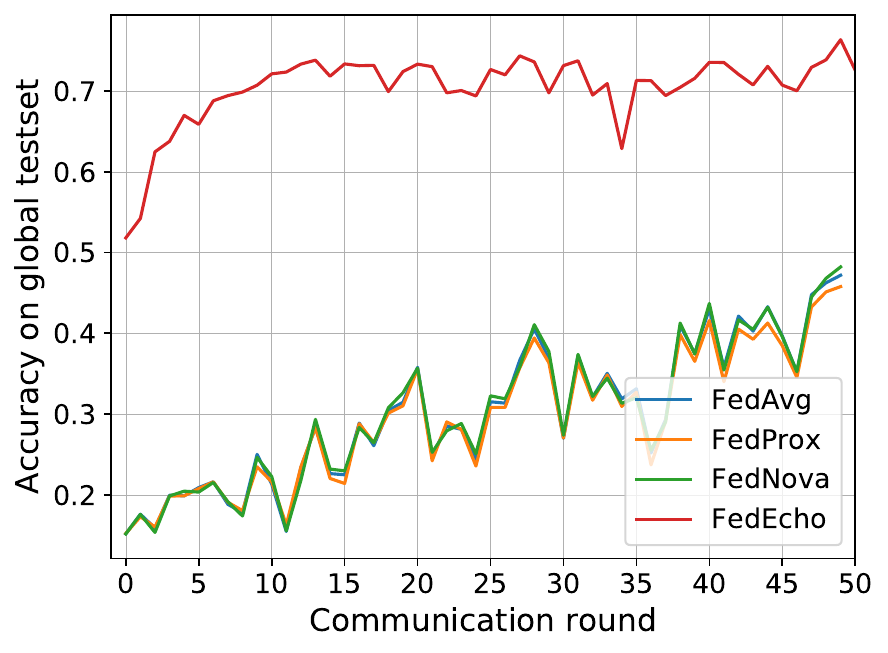}
}%

\subfloat[MNIST,$\frac{m}{N}$=$\frac{10}{100}$,$\#Class$=$2$]{
\includegraphics[width=0.38\linewidth]{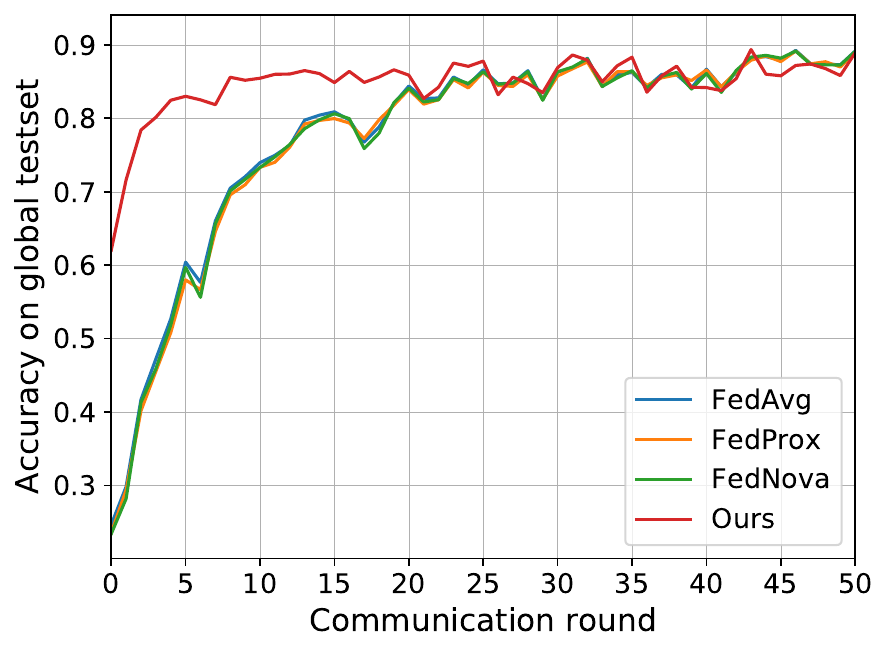}
}%
\subfloat[FEMNIST,$\frac{m}{N}$=$\frac{20}{1000}$,$\#Class$=$2$]{
\includegraphics[width=0.38\linewidth]{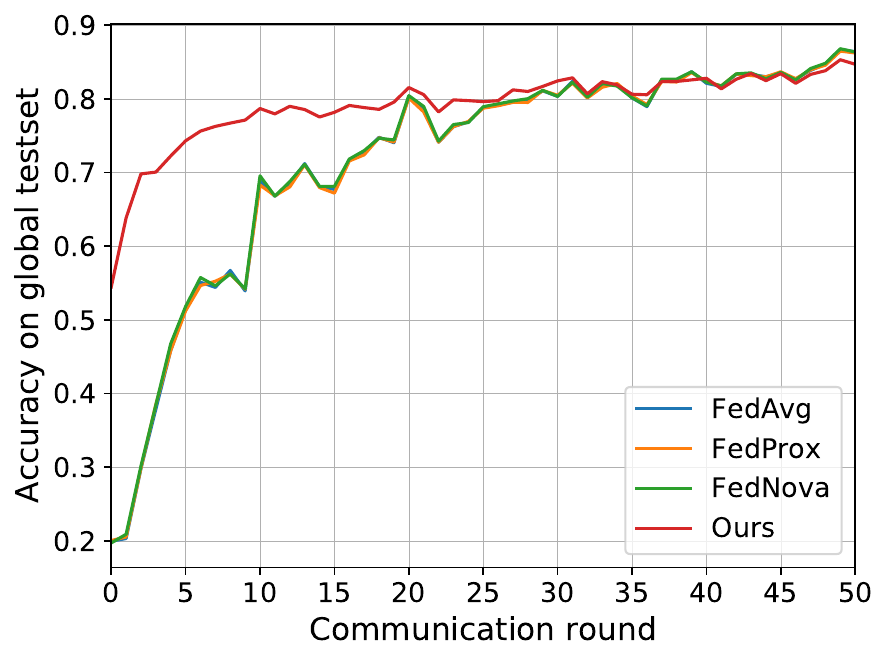}
}%
\caption{Performance comparison on the MNIST, FEMNIST data sets. Sample $m$ clients (local models) from $N$ clients in each communication round. $\#Class=2$ means that each client has $2$ classes of data in the Non-IID setting. When the data category distributions of the clients differ greatly, our method converges faster than other compared methods. Better viewed in color.}
\label{label2}
\end{figure*}

\section{Experiments}
\label{experiments}

To validate the effectiveness of  MA-Echo, we first investigate model aggregation of one-shot FL in three scenarios: (1) aggregating MLPs  (four fully-connected layers: $784\rightarrow 400 \rightarrow 200 \rightarrow 100 \rightarrow 10$) on the MNIST handwritten digits dataset~\cite{lecun-mnisthandwrittendigit-2010} ; (2) aggregating CNNs (three convolution layers and three fully-connected layers) on the CIFAR-10 dataset~\cite{krizhevsky2009learning}; and (3) aggregating the decoders of conditional variational auto-encoders (CVAE)~\cite{NIPS2015_8d55a249}  (the decoder has three fully-connected layers: $30\rightarrow 256 \rightarrow 512\rightarrow 784$) on MNIST.

Following the existing work \cite{YurochkinAGGHK19}, we sample $\mathbf{p}_{c}\sim\operatorname{Dir}(\beta\mathbf{1}_K)$ and allocate a $p_{c,k}$ proportion of the instances with label $c$ to the training set of the $k$-th local model. If $\beta$ is smaller, the label partition is more unbalanced; if $\beta$ approaches to infinity, all clients tend to have the identical label distribution over the training data, the effect of $\beta$ is illustrated in Figure~\ref{partition}. {We notice that when $\beta$ approaches to 0, there is almost no class overlap between different clients, which is very similar to the existing setting of federated partially supervised learning~\cite{dong2022federated}. However, in this paper, we do not specifically discuss this setting and instead validate our method in multiple general scenarios (i.e., $\beta>0$).} The ensemble of the local models is used to be the performance goal for model aggregation, which retains the knowledge of each local model to a large extent. 

In the experiments below, we first verify the effect of MA-Echo in one-shot federated learning, and then visualize the changes in model parameters during iterations. Second, we verify the influence of non-iid degree, local update size, parameter initialization and penalty coefficient $\mu$ on the performance of the aggregation. Finally, we put MA-Echo in multi-rounds federated learning to test its performance. To ensure the fairness of the comparison, we repeat the experiments three times with different random seeds and finally report the average of the three results. 

\subsection{One-shot performance}

Note that whether the local models have the same initial parameters before training seriously affects the aggregation result, a different initialization will significantly increase the difficulty of aggregation~\cite{McMahanMRHA17}. The reason is that, in the case of different initialization, the distance between any pair of local models will be larger. To comprehensively test the aggregation methods, we conduct experiments with both the same and different parameter initializations. Each local model is trained ten epochs in the training stage using SGD optimizer with an initial learning rate of 0.01 and momentum of 0.5.

We consider the aggregation of 2 and 5 local models in the extreme non-identical distribution scenarios where $\beta=0.01$. Aggregated results for a larger number of models will be shown in the multi-rounds experiment.  See the result in Figure~\ref{agr_exp}, MA-Echo outperforms the other methods, especially in the different-initialization setting. We can also see that for CNNs aggregation, the effectiveness of MA-Echo+OT is evident in the different-initialization setting. The reason may be that the permutation matrix provides global model parameters a good initial iteration value for MA-Echo.

{To demonstrate the scalability of MA-Echo, we conduct a multi-model aggregation experiment (up to 50 models). The model is a four layers MLP (the number of hidden layer neurons is 400, 200, 100, which is consistent with the MLP net in OTFusion[18]).  In order to show the time complexity of different methods, we also record the elapsed time of different methods. In Table~\ref{table_mt},  `Local acc' is the average performance of the local models on global testing data (without FL at all). It can be seen that 1) The performance of DENSE is significantly improved compared with other baseline methods, however, DENSE distills based on the ensemble model, it is difficult to exceed the ensemble performance. In our method, the projection matrix is introduced for assistance, which can significantly improve the accuracy of aggregated model and even exceed model ensemble. 2) When the number of models increases, our method still maintains leading performance. 3) Due to the training of generators and multi-teacher distillation, DENSE takes significantly more time than other methods.}

{For more complex data sets, such as CIFAR100, we use the pre-trained ResNet18 to fine tune and aggregate the parts of tuned parameters (two fully-connected layers) on CIFAR100. The results are in Table~\ref{table_c100}, for the aggregation of tuned parameters, our method still maintains significant performance improvement.}

We also train two CVAEs, the one is trained by $\{0,1,3,4,7\}$ categories, the other one is trained by $\{2,5,6,8,9\}$ categories. Then we compare the images generated by five different decoders: (1) Model1: the decoder from the first model; (2) Model2: the decoder from the second model; (3) Average: average the two trained decoders; (4) Ours: the decoder aggregated by MA-Echo (5)GT decoder: a model trained by the whole MNIST dataset. As shown in Figure~\ref{cvae}, the two local models can only generate the learned digits during training procedure. Our aggregation model can generate all categories of digits, and even close to the images of GT model.

{In the above experiments, we show the data heterogeneity caused by class imbalance. To verify the effect of MA-Echo under data heterogeneity caused by domain feature shift, we also conduct the other experiments. we use FEMNIST and DomainNet~\cite{peng2019moment} datasets. FEMNIST contains handwritten digital data labeled by more than 3000 users. We allocate data according to user IDs. Data from different clients come from different users. User style leads to domain feature shift among clients. DomainNet is a large-scale multi-domain dataset, including six domains with different styles: \textit{clipart}, \textit{infograph}, \textit{painting}, \textit{quickdraw}, \textit{real}, and \textit{sketch}. Each domain contains 345 classes. For FEMNIST data, we still use the MLP net. For DomainNet, we use the pre-trained ResNet34 model and freeze the feature extraction part, then we add two learnable fully-connected layers (including one relu layer) at the end of the model. Finally, we aggregate the two fully-connected layers. Table~\ref{table_ds} shows the aggregation effect under the domain feature shift setting. Due to the small differences among domains, the performance of various methods of FEMNIST is similar. For DomainNet, due to the large differences among domains, the effect of directly averaging is very poor. As a result, MA-Echo can greatly improve the aggregation performance.
}

\subsection{Visualization  }
In Eq.\ref{optv}, we hope $\{\mathbf{v}_i\}_{i=1}^N$ to be close to $\mathbf{w}$; at the same time, $\mathbf{v}_i$ should avoid forgetting the knowledge learned by $\mathbf{w}_i$. To verify these two purposes, we use t-sne~\cite{maaten2008visualizing} to visualize the matching of the third layer (the third layer in MLP, which has one hundred 200-dimensional parameter vectors) of each model in three-MLP aggregation. In Figure \ref{t_sne}, each number represents a parameter vector, and each color corresponds to a model. Compared with the original model, as the iteration progresses, these layers' parameter vectors reach a perfect match.

To visualize the optimization trajectory of $W^{l(t)}$ in the parameter space in three-MLPs aggregation, we use an indirect method~\cite{garipov2018loss,visualloss} to visualize the 2D-loss surface. Flatten the complete model parameters, and then we can get a vector for each model. For each step, repeat the flatten operation, we get $\{g^0,\cdots,g^\tau\}$, where $g^0 = \text{Flatten}(\{W^{1(0)},\cdots,W^{L(0)}\})$, then use $u=g^1-g^0$ and $u=g^2-g^0$ to form an orthogonal basis $\hat{u},\hat{v}$ in the 2D plane. The model corresponding to each point on the 2D plane is $P(x,y)=g^0+x\hat{u}+y\hat{v}$. Project the remaining $g^t$ into this plane, and calculate the loss value of each point in the test set, we can get the loss surface in Figure~\ref{sur}. As expected, MA-Echo is exploring the lower loss area in the parameter space during the iterations.

\subsection{The influence of different settings.}
\textbf{The influence of non-iid degree.} We conduct more experiments for different $\beta$ in two-MLPs aggregation on MNIST and two-CNNs aggregation on CIFAR-10. Since PFNM does not work for CNNs, we compare with FedMA instead in two-CNNs aggregation. The comparison results are reported in Table \ref{table_noniid}, MA-Echo achieves the best results in most cases and in different degrees of non-identicalness; OT is greatly improved when combined with MA-Echo. 

\textbf{The influence of local update.}
In the above experiments, we let the local model train for $10$ epochs. Here we try to observe the aggregation effect of the local model without sufficient training. We train the local models for different SGD steps. Note that in this case, 156 steps$\approx$ 1 epoch. We do 5-MLPs aggregation experiment, the result in Table~\ref{table_local} shows that the quality of the local model training is positively correlated with the aggregation effect, and under different settings, MA-Echo shows better performance than other methods.

\textbf{The influence of different initializations.}
We also test the effect of different initializations of $W^{l(0)}$ in Algorithm~\ref{algo1} on the performance. 
For the parameters of the classification layer, we still adopt the averaging operation; for the other layers, we choose three different initialization strategies, as shown in Figure~\ref{diff}. We split the MNIST data set for $5$ local models with $\beta=0.01$, then aggregate all the local models. The average strategy is still a good choice because it contains prior knowledge of multiple local models. In addition, we can see when we use the random strategy, MA-Echo can also increase its performance. When we use one local model for initialization, the accuracy fluctuates back and forth. However, no matter what kind of strategy is selected, MA-Echo will converge to similar performance. 

\textbf{The influence of Penalty coefficient $\mu$ in Eq.\ref{optv}.}
We do a comparative experiment to investigate the influence of the penalty coefficient $\mu$. From Eq.\ref{optv}, a smaller penalty coefficient $\mu$ will lead to more performance degradation for the local model. So we set $\mu=1$ and $\mu=200$ to investigate the difference between the accuracy of the local model and its initial model during iterations. As shown in Figure \ref{penltyC}, a smaller $\mu$ can bring appropriate relaxation to the projection constraint to improve the accuracy of the aggregated global model, at a cost of slight performance loss on local models. The reason is that, slightly losing the performance of local models may provide a larger search space for them to reach each other closer during optimization iterations.

\textbf{The influence of normalization.}
We use $\text{Norm}(\cdot)$ for diff-init CNN aggregation, here we respectively show the impact of whether normalization is used on the aggregation results. For the convenience of comparison, in aggregation experiment with $\text{Norm}(\cdot)$, we record the accuracy rate once every 10 iterations. One can see that in Figure~\ref{norm}, when we use normalization, the growth of aggregation effect shows an S-shaped trend and in the CNNs aggregation experiment, there is a better aggregation accuracy.

{
\textbf{The SVD decomposition for  $P$.}
 We conduct an experiment to show that we can easily reduce the size of the projection matrix by using the SVD decomposition for  $P^l$ without affecting the performance. We aggregate 20 MLP nets with $\beta=0.5$ in MNIST. The size of the original projection matrices are $784\times784$, $400 \times400$, $200 \times200$, $100 \times100$. We perform SVD decomposition on these matrices, and retain only a part of the principal components. then we use these principal components to restore the projection matrices on the server side. The results are in Table~\ref{table_svd}. When the matrices are compressed 100 times, the aggregation algorithm still retains 99\% of the performance. When the matrix is compressed 800 times, MA Echo still retains 94\% of the performance.
}

\subsection{Applied to Multi-rounds Federated Learning}
Considering that FL needs to allocate a large number of clients, we add the FEMNIST~\cite{caldas2018leaf} data set (we use the subset of FEMNIST, including $10$ labels and 382,705 handwritten digit images).  We consider $N$ clients, each of which has a 4-layer fully connected network as the local model, and sample $m$ clients for training in each round of communication. For FEMNIST, we consider $1000$ clients and sample $20$ for training. As adopted in many existing works, we construct the Non-IID data sets by randomly assigning $n$ labels to each client and then dividing the complete training set to different clients according to the labels.  
 
All local models are trained $10$ epochs by SGD optimizer with momentum $0.5$ and learning rate $0.01$. The regularization coefficient in FedProx is set to 0.1. All methods are implemented in PyTorch, and the code for the compared methods comes from an open-source repository~\cite{li2021federated}.

As shown in Figure~\ref{label2}, MA-Echo can significantly improve the global model in the first few rounds of communication and can achieve similar performance with much fewer communication rounds compared to other methods. As expected, MA-Echo tries to remember more knowledge learned from different clients -- this ability can help improve the aggregation efficiency to reduce the number of communication rounds.

\section{Conclusion}
In this paper, we focus on a new one-shot federated learning setting and proposes a novel model aggregation method named MA-Echo under this setting, which explores the common optima for all local models. Current approaches based on neuron-matching only permute rows of the parameter matrix without further changing their values. Thus, they can hardly achieve a good global model because the loss value of the averaging parameters is random. Motivated by this and inspired by continuous learning, we use the projection matrix as a kind of auxiliary information and then keep the original loss for each local model from being destroyed in aggregation. We demonstrate the excellent aggregation performance of MA-Echo through a large number of experiments. The experimental results have validated that the proposed method indeed can search for a lower loss global model. MA-Echo is also robust to models with varying degrees of non-identicalness of training data. In our future work, we will continue to improve MA-Echo to make it work in more complex neural networks.

\section*{Acknowledgement}
This work was supported in part by the National Natural Science Foundation of China (No.62176061), National Key R\&D Program of China (No.2021ZD0112803), STCSM projects (No.20511100400,  No.22511105000), the Shanghai Research and Innovation Functional Program (No.17DZ2260900), and the Program for Professor of Special Appointment (Eastern Scholar) at Shanghai Institutions of Higher Learning.



\section*{References}

\bibliography{refrence}

\begin{thebibliography}{10}
\expandafter\ifx\csname url\endcsname\relax
  \def\url#1{\texttt{#1}}\fi
\expandafter\ifx\csname urlprefix\endcsname\relax\def\urlprefix{URL }\fi
\expandafter\ifx\csname href\endcsname\relax
  \def\href#1#2{#2} \def\path#1{#1}\fi

\bibitem{krizhevsky2012imagenet}
A.~Krizhevsky, I.~Sutskever, G.~E. Hinton, Imagenet classification with deep
  convolutional neural networks, in: Advances in neural information processing
  systems, 2012, pp. 1097--1105.

\bibitem{he2016deep}
K.~He, X.~Zhang, S.~Ren, J.~Sun, Deep residual learning for image recognition,
  in: Proceedings of the IEEE conference on computer vision and pattern
  recognition, 2016, pp. 770--778.

\bibitem{kim2014convolutional}
Y.~Kim, \href{http://aclweb.org/anthology/D/D14/D14-1181.pdf}{Convolutional
  neural networks for sentence classification}, in: Proceedings of the 2014
  Conference on Empirical Methods in Natural Language Processing, {EMNLP} 2014,
  October 25-29, 2014, Doha, Qatar, {A} meeting of SIGDAT, a Special Interest
  Group of the {ACL}, 2014, pp. 1746--1751.
\newline\urlprefix\url{http://aclweb.org/anthology/D/D14/D14-1181.pdf}

\bibitem{McMahanMRHA17}
B.~McMahan, E.~Moore, D.~Ramage, S.~Hampson, B.~A. y~Arcas,
  \href{http://dblp.uni-trier.de/db/conf/aistats/aistats2017.html#McMahanMRHA17}{Communication-efficient
  learning of deep networks from decentralized data.}, in: A.~Singh, X.~J. Zhu
  (Eds.), AISTATS, Vol.~54 of Proceedings of Machine Learning Research, PMLR,
  2017, pp. 1273--1282.
\newline\urlprefix\url{http://dblp.uni-trier.de/db/conf/aistats/aistats2017.html#McMahanMRHA17}

\bibitem{yang2019federated}
Q.~Yang, Y.~Liu, T.~Chen, Y.~Tong, Federated machine learning: Concept and
  applications, ACM Transactions on Intelligent Systems and Technology (TIST)
  10~(2) (2019) 1--19.

\bibitem{chen2019communication}
Y.~Chen, X.~Sun, Y.~Jin, Communication-efficient federated deep learning with
  layerwise asynchronous model update and temporally weighted aggregation, IEEE
  transactions on neural networks and learning systems 31~(10) (2019)
  4229--4238.

\bibitem{sattler2019robust}
F.~Sattler, S.~Wiedemann, K.-R. M{\"u}ller, W.~Samek, Robust and
  communication-efficient federated learning from non-iid data, IEEE
  transactions on neural networks and learning systems 31~(9) (2019)
  3400--3413.

\bibitem{9463409}
B.~Gu, A.~Xu, Z.~Huo, C.~Deng, H.~Huang, Privacy-preserving asynchronous
  vertical federated learning algorithms for multiparty collaborative learning,
  IEEE Transactions on Neural Networks and Learning Systems (2021) 1--13\href
  {http://dx.doi.org/10.1109/TNNLS.2021.3072238}
  {\path{doi:10.1109/TNNLS.2021.3072238}}.

\bibitem{9632275}
F.~Sattler, T.~Korjakow, R.~Rischke, W.~Samek, Fedaux: Leveraging unlabeled
  auxiliary data in federated learning, IEEE Transactions on Neural Networks
  and Learning Systems (2021) 1--13\href
  {http://dx.doi.org/10.1109/TNNLS.2021.3129371}
  {\path{doi:10.1109/TNNLS.2021.3129371}}.

\bibitem{guha2019one}
N.~Guha, A.~Talwalkar, V.~Smith, One-shot federated learning, arXiv preprint
  arXiv:1902.11175.

\bibitem{salehkaleybar2021one}
S.~Salehkaleybar, A.~Sharifnassab, S.~J. Golestani, One-shot federated
  learning: theoretical limits and algorithms to achieve them, Journal of
  Machine Learning Research 22~(189) (2021) 1--47.

\bibitem{DBLP:conf/ijcai/LiHS21}
Q.~Li, B.~He, D.~Song, \href{https://doi.org/10.24963/ijcai.2021/205}{Practical
  one-shot federated learning for cross-silo setting}, in: Z.~Zhou (Ed.),
  Proceedings of the Thirtieth International Joint Conference on Artificial
  Intelligence, {IJCAI} 2021, Virtual Event / Montreal, Canada, 19-27 August
  2021, ijcai.org, 2021, pp. 1484--1490.
\newblock \href {http://dx.doi.org/10.24963/ijcai.2021/205}
  {\path{doi:10.24963/ijcai.2021/205}}.
\newline\urlprefix\url{https://doi.org/10.24963/ijcai.2021/205}

\bibitem{zhang2021fedzkt}
L.~Zhang, X.~Yuan, Fedzkt: Zero-shot knowledge transfer towards heterogeneous
  on-device models in federated learning, arXiv preprint arXiv:2109.03775.

\bibitem{zhou2020distilled}
Y.~Zhou, G.~Pu, X.~Ma, X.~Li, D.~Wu, Distilled one-shot federated learning,
  arXiv preprint arXiv:2009.07999.

\bibitem{hinton2015distilling}
G.~Hinton, O.~Vinyals, J.~Dean, Distilling the knowledge in a neural network,
  arXiv preprint arXiv:1503.02531.

\bibitem{wang2018dataset}
T.~Wang, J.-Y. Zhu, A.~Torralba, A.~A. Efros, Dataset distillation, arXiv
  preprint arXiv:1811.10959.

\bibitem{zhangdense}
J.~Zhang, C.~Chen, B.~Li, L.~Lyu, S.~Wu, S.~Ding, C.~Shen, C.~Wu, Dense:
  Data-free one-shot federated learning, in: Advances in Neural Information
  Processing Systems 2022.

\bibitem{YurochkinAGGHK19}
M.~Yurochkin, M.~Agarwal, S.~Ghosh, K.~H. Greenewald, T.~N. Hoang, Y.~Khazaeni,
  \href{http://dblp.uni-trier.de/db/conf/icml/icml2019.html#YurochkinAGGHK19}{Bayesian
  nonparametric federated learning of neural networks.}, in: K.~Chaudhuri,
  R.~Salakhutdinov (Eds.), ICML, Vol.~97 of Proceedings of Machine Learning
  Research, PMLR, 2019, pp. 7252--7261.
\newline\urlprefix\url{http://dblp.uni-trier.de/db/conf/icml/icml2019.html#YurochkinAGGHK19}

\bibitem{singh2020model}
S.~P. Singh, M.~Jaggi, Model fusion via optimal transport, Advances in Neural
  Information Processing Systems 33.

\bibitem{WangYSPK20}
H.~Wang, M.~Yurochkin, Y.~Sun, D.~S. Papailiopoulos, Y.~Khazaeni, Federated
  learning with matched averaging., in: International Conference on Learning
  Representations, 2020.

\bibitem{NIPS2015_8d55a249}
K.~Sohn, H.~Lee, X.~Yan,
  \href{https://proceedings.neurips.cc/paper/2015/file/8d55a249e6baa5c06772297520da2051-Paper.pdf}{Learning
  structured output representation using deep conditional generative models},
  in: C.~Cortes, N.~Lawrence, D.~Lee, M.~Sugiyama, R.~Garnett (Eds.), Advances
  in Neural Information Processing Systems, Vol.~28, Curran Associates, Inc.,
  2015.
\newline\urlprefix\url{https://proceedings.neurips.cc/paper/2015/file/8d55a249e6baa5c06772297520da2051-Paper.pdf}

\bibitem{konevcny2016federated}
J.~Kone{\v{c}}n{\`y}, H.~B. McMahan, F.~X. Yu, P.~Richt{\'a}rik, A.~T. Suresh,
  D.~Bacon, Federated learning: Strategies for improving communication
  efficiency, arXiv preprint arXiv:1610.05492.

\bibitem{conf/mlsys/LiSZSTS20}
T.~Li, A.~K. Sahu, M.~Zaheer, M.~Sanjabi, A.~Talwalkar, V.~Smith,
  \href{http://dblp.uni-trier.de/db/conf/mlsys/mlsys2020.html#LiSZSTS20}{Federated
  optimization in heterogeneous networks.}, in: I.~S. Dhillon, D.~S.
  Papailiopoulos, V.~Sze (Eds.), MLSys, mlsys.org, 2020.
\newline\urlprefix\url{http://dblp.uni-trier.de/db/conf/mlsys/mlsys2020.html#LiSZSTS20}

\bibitem{acar2021federated}
D.~A.~E. Acar, Y.~Zhao, R.~M. Navarro, M.~Mattina, P.~N. Whatmough,
  V.~Saligrama, Federated learning based on dynamic regularization, in:
  International Conference on Learning Representations, 2021.

\bibitem{li2021fedbn}
X.~Li, M.~Jiang, X.~Zhang, M.~Kamp, Q.~Dou, Fedbn: Federated learning on
  non-iid features via local batch normalization, ICLR.

\bibitem{karimireddy2020scaffold}
S.~P. Karimireddy, S.~Kale, M.~Mohri, S.~Reddi, S.~Stich, A.~T. Suresh,
  Scaffold: Stochastic controlled averaging for federated learning, in:
  International Conference on Machine Learning, PMLR, 2020, pp. 5132--5143.

\bibitem{reddi2020adaptive}
S.~Reddi, Z.~Charles, M.~Zaheer, Z.~Garrett, K.~Rush, J.~Kone{\v{c}}n{\`y},
  S.~Kumar, H.~B. McMahan, Adaptive federated optimization, arXiv preprint
  arXiv:2003.00295.

\bibitem{wang2020tackling}
J.~Wang, Q.~Liu, H.~Liang, G.~Joshi, H.~V. Poor, Tackling the objective
  inconsistency problem in heterogeneous federated optimization, Advances in
  Neural Information Processing Systems 33.

\bibitem{zhu2021data}
Z.~Zhu, J.~Hong, J.~Zhou, Data-free knowledge distillation for heterogeneous
  federated learning, in: International Conference on Machine Learning, PMLR,
  2021, pp. 12878--12889.

\bibitem{yoonfedmix}
T.~Yoon, S.~Shin, S.~J. Hwang, E.~Yang, Fedmix: Approximation of mixup under
  mean augmented federated learning, International Conference on Learning
  Representations.

\bibitem{mohri2019agnostic}
M.~Mohri, G.~Sivek, A.~T. Suresh, Agnostic federated learning, in:
  International Conference on Machine Learning, PMLR, 2019, pp. 4615--4625.

\bibitem{li2019fair}
T.~Li, M.~Sanjabi, A.~Beirami, V.~Smith, Fair resource allocation in federated
  learning, International Conference on Learning Representations.

\bibitem{hu2020fedmgda+}
Z.~Hu, K.~Shaloudegi, G.~Zhang, Y.~Yu, Fedmgda+: Federated learning meets
  multi-objective optimization, arXiv preprint arXiv:2006.11489.

\bibitem{lin2020ensemble}
T.~Lin, L.~Kong, S.~U. Stich, M.~Jaggi, Ensemble distillation for robust model
  fusion in federated learning, Advances in Neural Information Processing
  Systems 33.

\bibitem{li2021model}
Q.~Li, B.~He, D.~Song, Model-contrastive federated learning, Proceedings of the
  IEEE/CVF Conference on Computer Vision and Pattern Recognition.

\bibitem{chen2020simple}
T.~Chen, S.~Kornblith, M.~Norouzi, G.~Hinton, A simple framework for
  contrastive learning of visual representations, in: International conference
  on machine learning, PMLR, 2020, pp. 1597--1607.

\bibitem{gou2021knowledge}
J.~Gou, B.~Yu, S.~J. Maybank, D.~Tao, Knowledge distillation: A survey,
  International Journal of Computer Vision 129~(6) (2021) 1789--1819.

\bibitem{shin2020xor}
M.~Shin, C.~Hwang, J.~Kim, J.~Park, M.~Bennis, S.-L. Kim, Xor mixup:
  Privacy-preserving data augmentation for one-shot federated learning, arXiv
  preprint arXiv:2006.05148.

\bibitem{journals/siamma/AguehC11}
M.~Agueh, G.~Carlier,
  \href{http://dblp.uni-trier.de/db/journals/siamma/siamma43.html#AguehC11}{Barycenters
  in the wasserstein space.}, SIAM J. Math. Analysis 43~(2) (2011) 904--924.
\newline\urlprefix\url{http://dblp.uni-trier.de/db/journals/siamma/siamma43.html#AguehC11}

\bibitem{zeng2019continual}
G.~Zeng, Y.~Chen, B.~Cui, S.~Yu, Continual learning of context-dependent
  processing in neural networks, Nature Machine Intelligence 1~(8) (2019)
  364--372.

\bibitem{farajtabar2020orthogonal}
M.~Farajtabar, N.~Azizan, A.~Mott, A.~Li, Orthogonal gradient descent for
  continual learning, in: International Conference on Artificial Intelligence
  and Statistics, PMLR, 2020, pp. 3762--3773.

\bibitem{ChaudhryOrthogSubspaceCL}
A.~Chaudhry, N.~Khan, P.~K. Dokania, P.~H. Torr, Continual learning in low-rank
  orthogonal subspaces, in: NeurIPS, 2020.

\bibitem{fliege2000steepest}
J.~Fliege, B.~F. Svaiter, Steepest descent methods for multicriteria
  optimization, Mathematical Methods of Operations Research 51~(3) (2000)
  479--494.

\bibitem{andersen2013cvxopt}
M.~S. Andersen, J.~Dahl, L.~Vandenberghe, Cvxopt: A python package for convex
  optimization, abel. ee. ucla. edu/cvxopt 88.

\bibitem{lecun-mnisthandwrittendigit-2010}
Y.~LeCun, C.~Cortes, \href{http://yann.lecun.com/exdb/mnist/}{{MNIST}
  handwritten digit database}.
\newline\urlprefix\url{http://yann.lecun.com/exdb/mnist/}

\bibitem{krizhevsky2009learning}
A.~Krizhevsky,
  \href{https://www.cs.toronto.edu/~kriz/learning-features-2009-TR.pdf}{Learning
  multiple layers of features from tiny images} (2009) 32--33.
\newline\urlprefix\url{https://www.cs.toronto.edu/~kriz/learning-features-2009-TR.pdf}

\bibitem{dong2022federated}
N.~Dong, M.~Kampffmeyer, I.~Voiculescu, E.~Xing, Federated partially supervised
  learning with limited decentralized medical images, IEEE Transactions on
  Medical Imaging.

\bibitem{peng2019moment}
X.~Peng, Q.~Bai, X.~Xia, Z.~Huang, K.~Saenko, B.~Wang, Moment matching for
  multi-source domain adaptation, in: Proceedings of the IEEE/CVF international
  conference on computer vision, 2019, pp. 1406--1415.

\bibitem{maaten2008visualizing}
L.~v.~d. Maaten, G.~Hinton, Visualizing data using t-sne, Journal of machine
  learning research 9~(Nov) (2008) 2579--2605.

\bibitem{garipov2018loss}
T.~Garipov, P.~Izmailov, D.~Podoprikhin, D.~Vetrov, A.~G. Wilson, Loss
  surfaces, mode connectivity, and fast ensembling of dnns, in: Proceedings of
  the 32nd International Conference on Neural Information Processing Systems,
  2018, pp. 8803--8812.

\bibitem{visualloss}
H.~Li, Z.~Xu, G.~Taylor, C.~Studer, T.~Goldstein, Visualizing the loss
  landscape of neural nets, in: Neural Information Processing Systems, 2018.

\bibitem{caldas2018leaf}
S.~Caldas, S.~M.~K. Duddu, P.~Wu, T.~Li, J.~Kone{\v{c}}n{\`y}, H.~B. McMahan,
  V.~Smith, A.~Talwalkar, Leaf: A benchmark for federated settings,
  International Conference on Machine Learning (ICML) Workshop on Federated
  Learning for Data Privacy and Conﬁdentiality.

\bibitem{li2021federated}
Q.~Li, Y.~Diao, Q.~Chen, B.~He, Federated learning on non-iid data silos: An
  experimental study, arXiv preprint arXiv:2102.02079.

\end{thebibliography}
\end{document}